\documentclass[runningheads]{llncs}

\usepackage{eccv}

\usepackage{eccvabbrv}

\usepackage{graphicx}
\usepackage{booktabs}
\usepackage{wrapfig}

\usepackage[accsupp]{axessibility}  

\usepackage{booktabs}      
\usepackage{multirow}      
\usepackage[table]{xcolor} 
\usepackage{float} 
\usepackage{tcolorbox}

\newcommand{\best}[1]{\textbf{#1}}

\makeatletter\renewcommand\paragraph{\@startsection{paragraph}{4}{\z@}
  {.5em \@plus1ex \@minus.2ex}{-.5em}{\normalfont\normalsize\bfseries}}\makeatother

\usepackage[hyperfootnotes=false]{hyperref}
\usepackage{marvosym}

\usepackage{orcidlink}

\newcommand{\blfootnote}[1]{%
  \begingroup
  \renewcommand{\thefootnote}{}%
  \footnotetext{#1}%
  \addtocounter{footnote}{-1}%
  \endgroup
}

\begin{document}

\title{Learning to Think Fast and Slow for Visual Language Models} 


\author{Chenyu Lin\inst{1} \and
Cheng Chi\inst{2}\textsuperscript{\Letter} \and
Jinlin Wu\inst{3} \and
Sharon Li\inst{4} \and
Kaiyang Zhou\inst{1}\textsuperscript{\Letter} }

\authorrunning{C. Lin et al.}

\institute{Hong Kong Baptist University\and 
Beijing Academy of Artificial Intelligence \and
Institute of Automation, CAS \and
University of Wisconsin-Madison
\\[0.6em]
\texttt{https://github.com/maifoundations/DualMindVLM}}

\maketitle

\blfootnote{\textsuperscript{\Letter} Corresponding authors.}

\begin{abstract}
When faced with complex problems, we tend to engage in slower, more deliberate thinking. In contrast, for simple questions we give quick, intuitive responses. This dual-system thinking approach allows us to allocate cognitive resources efficiently, reserving deeper analytical effort for tasks that truly require it. However, existing reasoning-oriented visual language models (VLMs) are mostly trained to generate uniformly long reasoning, leading to substantial token waste when concise answers would suffice. In this paper, we observe that pre-trained, general-purpose VLMs manifest variations in response length for different question types, e.g., longer reasoning for math questions while shorter on perception problems. Different from existing work that overrides this prior by stimulating long reasoning without considering the problem complexity, we propose to leverage this prior to develop an explicit dual-mode thinking mechanism. Specifically, we anchor each training instance to either a fast or slow thinking prefix consistent with the model's natural response length tendency. Then, GRPO is adapted to learning dual-system thinking, enabling both autonomous and manual thinking mode selection. Extensive experiments across a wide variety of visual reasoning benchmarks demonstrate that our model, named DualMindVLM, significantly outperforms the base model and achieves state-of-the-art reasoning performance while maintaining high token efficiency.
\end{abstract}

\section{Introduction}
\label{sec:intro}

Human cognition is widely recognized to operate through two thinking systems: System 1 and System 2~\cite{kahneman2011thinking, evans2013dual, evans2017dual}. System 1 enables fast, automatic responses to routine or simple scenarios, while System 2 engages in slow, deliberate reasoning for intricate or unknown challenges. Remarkably, humans dynamically switch between these two modes depending on task demands. Such adaptive allocation of reasoning effort offers valuable inspiration for designing more cognitively-aligned visual language models (VLMs).

\begin{figure*}[t]
    \centering
    \includegraphics[width=\textwidth]{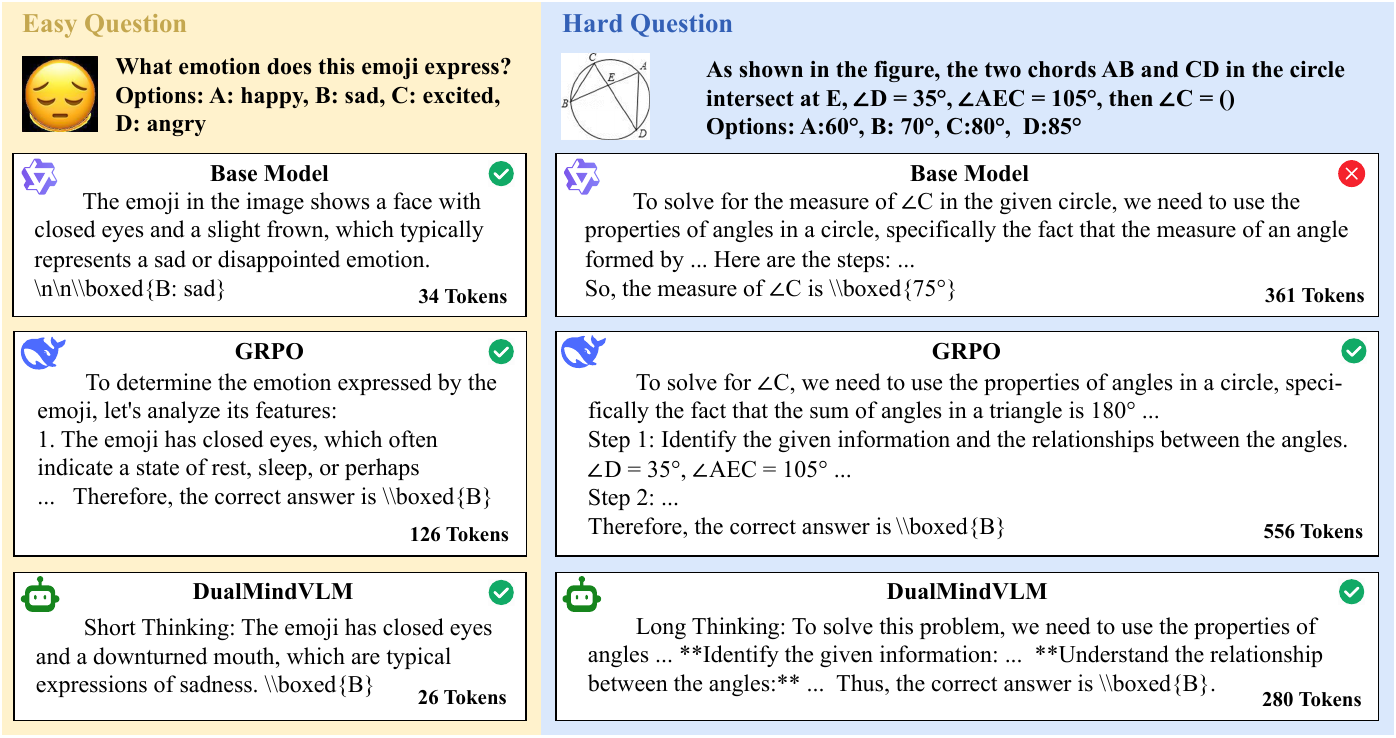}
    \caption{\textbf{Comparison among the base model, the GRPO model and our DualMindVLM}. For simple queries, the GRPO model tends to produce unnecessarily long responses, leading to additional computational overhead for questions that the base model can handle efficiently. In contrast, our model adaptively balances response length by maintaining concise answers when elaborate reasoning is unnecessary and engaging in more detailed reasoning when needed through two automatically selected thinking modes.}
    \label{fig:intro_case}
\end{figure*}


Current research on visual reasoning models primarily emphasizes step-by-step reasoning~\cite{yao2024mulberry, xu2025llava, dong2025insight, yang2025r1, zhang2025r1, deng2025openvlthinker, xia2025visionary, wang2025vl, xia2025bootstrapping, huang2025vision, shen2025vlm}, encouraging behaviors such as detailed image description or reflective reasoning to elongate their reasoning chains. However, existing approaches ignore the human-like dual-mode thinking mechanism, causing excessive reasoning on all queries and thus leading to redundant token usage. As shown in Fig.~\ref{fig:intro_case}, the model trained with Group Relative Policy Optimization (GRPO)~\cite{shao2024deepseekmath}, produces substantially longer reasoning chains compared to the base model. While such step-by-step reasoning benefits challenging problems like math (Fig.~\ref{fig:intro_case} right), it incurs unnecessary computational overhead on simpler ones (Fig.~\ref{fig:intro_case} left).

\begin{figure}[t]
    \centering
    \includegraphics[width=\columnwidth]{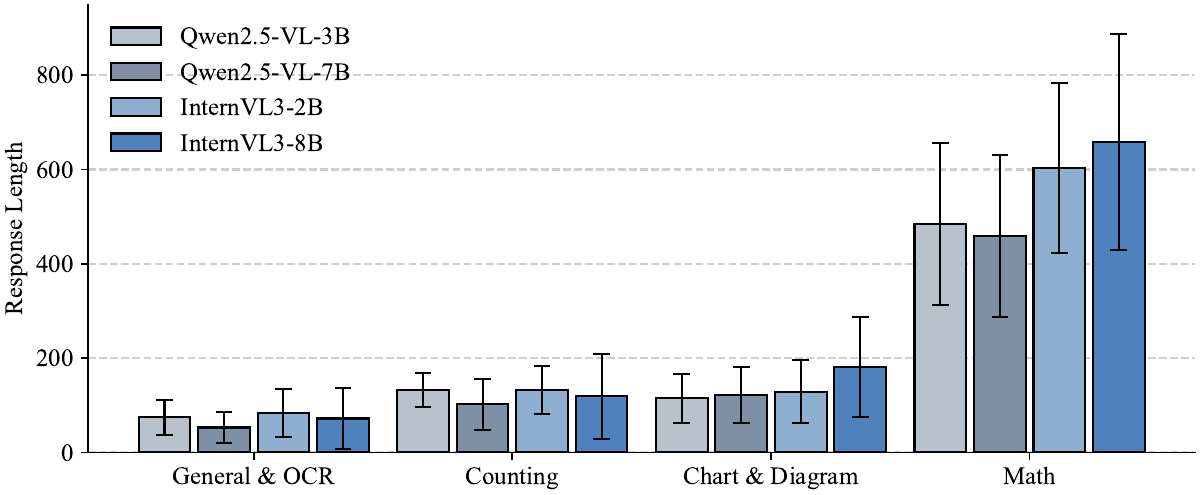}
    \caption{\textbf{Average response lengths of four pre-trained general-purpose VLMs across a variety of VQA tasks.} Response lengths remain relatively consistent within the same task while varying across tasks, suggesting the presence of an implicit response-length prior inherited from large-scale pre-training.}
    \label{fig:observation}
\end{figure}

\begin{wrapfigure}{r}{0.5\columnwidth}
    \vspace{-\intextsep}
    \centering
    \includegraphics[width=\linewidth]{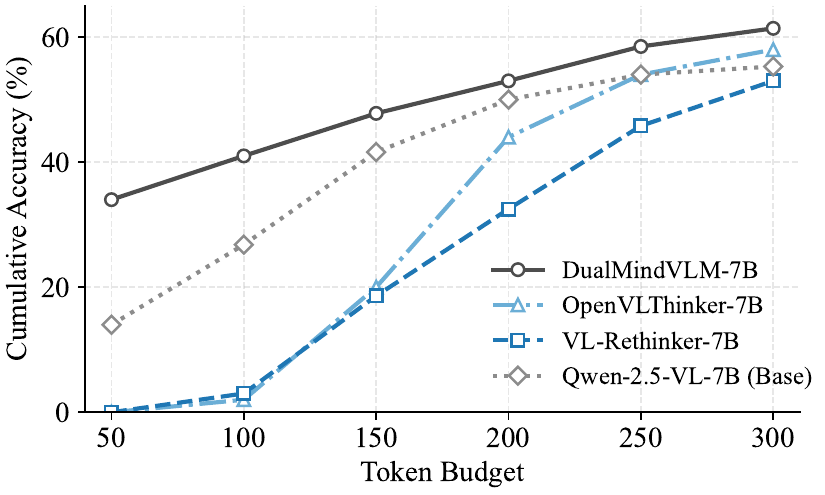}
    \caption{{Cumulative accuracy under varying token budgets on MMStar~\cite{chen2024we}}.}
    \label{fig:intro_budgets}
    \vspace{-\intextsep}
\end{wrapfigure}

This limitation raises a natural question: can VLMs adaptively switch between fast and slow thinking as humans do? To explore this possibility, we first examine the response length patterns of pre-trained VLMs. Even without explicit dual-mode training, these models manifest variations in response length across different tasks. As shown in Fig.~\ref{fig:observation}, we investigate four representative VLMs of different scales and architectures, including Qwen2.5-VL-3B/7B~\cite{bai2025qwen2} and InternVL3-2B/8B~\cite{zhu2025internvl3}. The results reveal an interesting pattern: the response lengths of different models are generally consistent within the same task while varying across tasks of different complexities. Specifically, these models consistently generate long responses on mathematical problems, medium-length responses on chart understanding and counting tasks, and short responses on general perception and OCR-related questions. This pattern suggests that an implicit prior on response length emerges during large-scale pre-training. However, most existing reasoning methods override this prior by uniformly encouraging extended reasoning. As a consequence, these methods lose adaptive thinking and rely on long output tokens to get the right answer. As evidenced in Fig.~\ref{fig:intro_budgets}, the two reasoning models, namely OpenVLThinker~\cite{deng2025openvlthinker} and VL-Rethinker~\cite{wang2025vl}, significantly underperform the base model with a limited token budget, such as 100 tokens, and start to show advantages when the token length is over 250.

In this work, we introduce \textbf{DualMindVLM}, a dual-mode thinking VLM that leverages the model's intrinsic prior on response length to develop two controllable thinking modes, enabling automatic switching between fast and slow thinking. The approach consists of two stages, as illustrated in Fig.~\ref{fig:framework}. In the first stage, each training instance is anchored to a thinking prefix following the model's natural response length tendency. In the second stage, we employ GRPO with partially-constrained rollouts where half of the trajectories are generated with a thinking mode prefix while the other half are freely generated. This design reinforces both thinking modes while allowing the model to autonomously select the appropriate prefix at inference time.

We conduct extensive experiments on a wide range of multimodal benchmarks spanning mathematics~\cite{lu2023mathvista, wang2024measuring}, science~\cite{kembhavi2016diagram, lu2022learn}, and general visual understanding problems~\cite{chen2024we, liu2024mmbench}. The results show that DualMindVLM consistently delivers highly competitive performance compared to state-of-the-art reasoning VLMs while maintaining high token efficiency.

In summary, our main contributions are threefold: 1) We identify an implicit prior on response length in pre-trained VLMs and show that it can be leveraged to develop an explicit dual-mode thinking mechanism. 2) We propose a two-stage training framework that stabilizes and strengthens both thinking modes while enabling automatic mode selection at inference time. 3) Extensive experiments are conducted on six multimodal benchmarks to demonstrate the effectiveness of DualMindVLM. Code and models will be made publicly available to facilitate future research.

\section{Related Work}
\label{sec:related}

\paragraph{Visual reasoning.}
Driven by the recent advances in reasoning capabilities of LLMs~\cite{jaech2024openai, guo2025deepseek, team2025kimi}, the vision community has increasingly focused on equipping VLMs with step-by-step reasoning abilities. Early approaches~\cite{yao2024mulberry, xu2025llava} rely on supervised fine-tuning (SFT) with curated chain-of-thought data to instill structured reasoning patterns. With the introduction of GRPO~\cite{shao2024deepseekmath}, subsequent work explores reinforcement learning (RL) methods that leverage verifiable rewards to further elicit deliberate reasoning. Many studies adopt a two-stage SFT+RL paradigm~\cite{deng2025openvlthinker, zhang2025r1, yang2025r1, tan2025reason, dong2025insight, huang2025vision, zhou2025roborefer}, where SFT initializes extended reasoning behaviors that are subsequently reinforced through RL. Others pursue RL-only strategies~\cite{xia2025visionary, wang2025vl, wang2025sota, meng2025mm}, eliciting longer trajectories via prompt guidance and emphasizing challenging samples, detailed descriptions, or reflective reasoning. Despite their effectiveness on complex tasks, these approaches uniformly encourage slow thinking, overlooking that not all queries require step-by-step deliberation and thus introducing unnecessary computational overhead on simpler ones.

\paragraph{Efficient reasoning.}
Improving the efficiency of reasoning models has been actively studied in the LLM domain. Chain-of-Draft~\cite{xu2025chain} encourages models to generate concise intermediate steps, while DAST~\cite{shen2025dast} and AdaCoT~\cite{lou2025adacot} employ RL to penalize unnecessarily long reasoning trajectories. Some approaches~\cite{agarwal2025gpt, yang2025qwen3} train unified models that support multiple thinking modes, yet users must still manually select the appropriate mode. In the multimodal domain, recent efforts have begun to explore efficiency-aware reasoning for VLMs. TON~\cite{wang2025think} enables exploration between thinking and non-thinking paths by explicitly removing intermediate reasoning traces, while FAST~\cite{xiao2025fast} incorporates response length as an optimization objective through tailored reward design. In contrast, our approach takes a different perspective: instead of optimizing length directly, we leverage the intrinsic length prior of pre-trained VLMs to explicitly separate fast and slow reasoning behaviors and jointly enhance them. This design avoids additional SFT stages or explicit length-based reward engineering, and results in controllable dual thinking modes with automatic mode switching.

\section{Methodology}

\begin{figure*}[t]
    \centering
    \includegraphics[width=\textwidth]{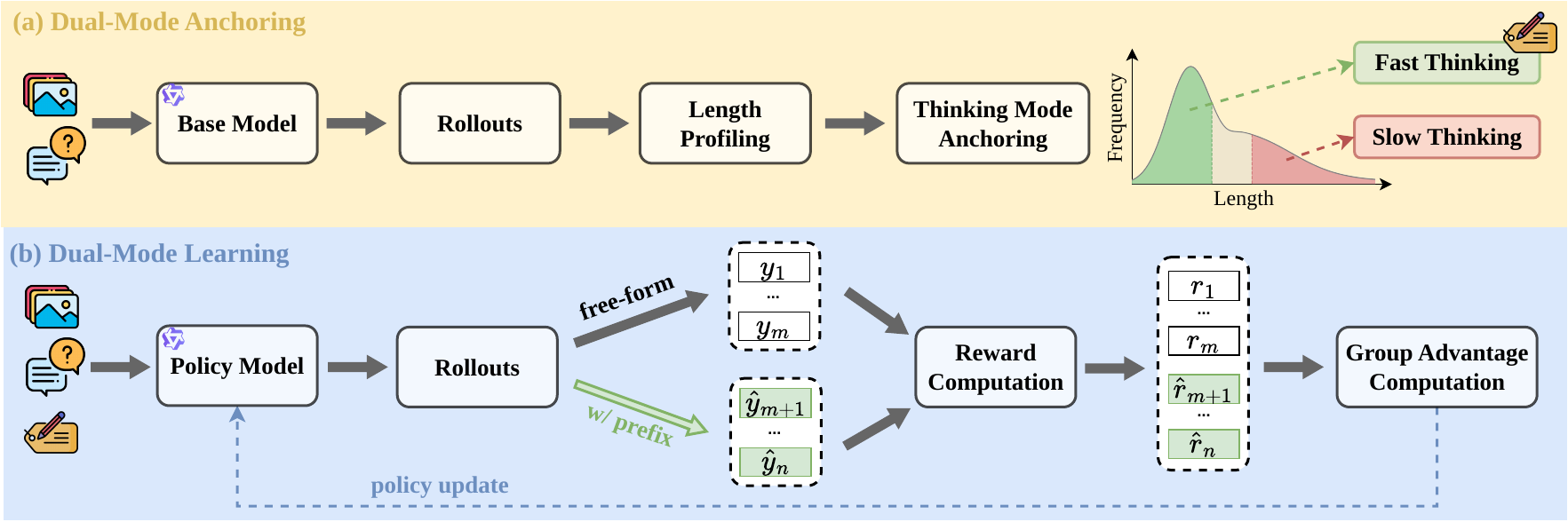}
    \caption{\textbf{Overview of DualMindVLM}. (a) For each VQA pair, we collect multiple rollouts from the base model and profile their response lengths. Based on the observed length tendency, we anchor the pair to a thinking mode by assigning the corresponding control prefix. (b) We then perform GRPO with partially constrained rollouts, where one group of candidates is generated using the assigned prefix, while the other is freely generated. A group-wise advantage is computed over all candidate responses to reinforce dual-mode reasoning behaviors and consistent prefix usage.}
    \label{fig:framework}
\end{figure*}

Existing visual reasoning methods primarily focus on System 2 thinking, i.e., generating detailed chain-of-thought reasoning, while overlooking the model's inherent length prior and the development of efficient System 1 thinking. As a result, they incur unnecessary token redundancy for simple queries. To fill the gap, we propose DualMindVLM, a dual-mode thinking model that is trained using RL and simple visual question-answer pairs.

As shown in Fig.~\ref{fig:framework}, the overall training pipeline of DualMindVLM consists of two stages. The first stage, \textit{dual-mode anchoring}, maps the model's implicit length prior to two thinking modes by partitioning the training data into two subsets. Instances associated with short responses are used to develop fast thinking, while those exhibiting long responses are used to develop slow thinking. Each thinking mode is associated with a control prefix that triggers the corresponding reasoning behavior. The second stage, \textit{dual-mode learning}, builds upon the anchored thinking modes to optimize reasoning in each mode and enable autonomous mode selection. Specifically, we employ GRPO with partially constrained rollouts, where one group is generated using the assigned control prefix, while the other is freely generated. The reward function is computed over both groups and favors responses that are correct and consistent with the anchored thinking mode, allowing the model to strengthen reasoning behavior and learn to use the appropriate prefix. Below we detail the designs of these two stages.

\begin{figure}[t]
\centering
\begin{tcolorbox}[
  colback=gray!10,
  colframe=gray!60!black,
  arc=3mm,
  boxrule=0.8pt,
  left=4pt,
  right=4pt,
  top=4pt,
  bottom=4pt,
  fonttitle=\bfseries,
  halign=left,
  title={System Prompt}
]
\small
You are a Vision-Language Model answering questions about images. 
Follow these rules strictly:\\
1. Judge the length of reasoning needed.\\
- Short: start with \texttt{"Short Thinking:"}.\\
- Long: start with \texttt{"Long Thinking:"}.\\
2. \texttt{"Short Thinking:"} give a concise thinking process which is sufficient to answer the question, then provide the final answer.\\
3. \texttt{"Long Thinking:"} give a structured reasoning process of the question and the image, including question analysis, visual details description, self-verification and then provide the final answer.\\
4. The final answer MUST BE put in \texttt{\textbackslash boxed\{\}}.
\end{tcolorbox}
\vspace{-10pt}
\caption{{Dual-mode system prompt for binding thinking modes with prefixes.}}
\label{fig:sys_prompt}
\end{figure}

\subsection{Dual-Mode Anchoring}
\label{sec:dual_mode_anchor}

Building upon our observation that pre-trained VLMs exhibit a systematic length prior—where response lengths tend to remain consistent within the same task while varying across tasks—we aim to convert this tendency into a controllable dual-mode thinking mechanism. This stage achieves two primary objectives: (1) anchoring data instances to distinct thinking modes based on the model's inherent length tendency, and (2) binding abstract thinking modes to explicit control prefixes to enable steerable reasoning.

\paragraph{Anchoring data to thinking modes.} We first anchor each training sample to a thinking mode consistent with the model's inherent length tendency. As illustrated in Fig.~\ref{fig:framework}(a), for each sample, we collect multiple rollouts (e.g., 8) from the base model to perform length profiling, where the average response length is computed as a proxy for the model's length tendency on that sample. To ensure the separation between fast and slow reasoning behaviors, we adopt straightforward length thresholds. A sample is anchored to fast thinking if the average length is below $\tau_{fast}$, or to slow thinking if it exceeds $\tau_{slow}$. Although simple, this procedure provides a clear signal for developing distinct thinking modes. We set $\tau_{fast}=100$ tokens and $\tau_{slow}=200$ tokens as the default thresholds. Sensitivity analysis for these thresholds is provided in Section~\ref{sec:futher_analysis}. To mitigate the problem of vanishing advantages~\cite{wang2025vl, meng2025mm}, we exclude samples with an average accuracy of 0 or 1. Such cases are unlikely to provide meaningful relative advantages for optimization.

\paragraph{Binding thinking modes to prefixes.}

To convert these abstract thinking modes into steerable behaviors, we bind each mode to a control prefix. As shown in Fig.~\ref{fig:sys_prompt}, we define a structured output format in which the model must generate a specific prefix before reasoning. Specifically, $p^{\text{fast}}=\texttt{"Short Thinking:"}$ is associated with concise processing, whereas $p^{\text{slow}}=\texttt{"Long Thinking:"}$ triggers detailed, multi-step reasoning. This prefix-to-mode binding establishes a steerable reasoning interface, translating previously anchored thinking modes into selectable behaviors. This design aligns the control prefixes with the model’s inherent response tendencies, making the prefixes a natural trigger for the corresponding thinking behaviors.

\subsection{Dual-Mode Learning}
\label{sec:dual_mode_learning}

The goal of this stage is to strengthen dual-mode reasoning and internalize prefix usage as an autonomous behavior of the model. We adopt GRPO~\cite{shao2024deepseekmath} with partially constrained rollouts: for each input, we sample a group of candidate responses, where half are generated with the anchored prefix and half are generated in free-form. See Fig.~\ref{fig:framework}(b) for illustration.

\paragraph{Hybrid group response sampling.}
Directly learning prefix usage from purely free-form rollouts suffers from a cold-start issue: the base model does not reliably emit the required prefixes or produces answers inconsistent with the chosen prefix. This problem leads to unstable training. To address this problem, we introduce hybrid group response sampling. Formally, given an input $x=(I,Q)$, we sample $n$ candidate responses from the sampling model $\pi_{\theta_{\text{old}}}$ and split them into two subgroups: a free-form subgroup $\{y_i\}_{i=1}^{m}$ and a prefix-conditioned subgroup $\{\hat{y}_i\}_{i=m+1}^{n}$. For each $\hat{y}_i$, we manually insert the anchored prefix $p^*$ before generation to enforce the desired output format. Below we discuss the reward and the training objective using only the notation $y$ for clarity.

Crucially, because the anchored prefix from the previous stage is aligned with the model's inherent length tendency, the prefix-conditioned subgroup naturally generate trajectories that are consistent with the intended behavior. These trajectories serve as stable behavioral anchors within the group, allowing the free-form subgroup to learn the mapping between prefixes and corresponding reasoning behaviors through group-wise comparisons. Once this mapping is established, the RL process can focus on increasing the likelihood of reasoning trajectories that lead to correct answers, thereby optimizing dual-mode thinking abilities.

\paragraph{Reward design.}
Each candidate response is scored by a joint reward:

\begin{equation}
    r(y_i)=r_a(y_i)+r_f(y_i)
\end{equation}

This reward is used to optimize answer correctness while promoting mode consistency.

\begin{itemize}
    \item Accuracy Reward ($r_a$): We set $r_a(y)=1$ if the predicted answer is correct and $0$ otherwise.
    \item Format Reward ($r_f$): To encourage prefix usage and prioritize consistency with the anchored mode, we define
    \begin{equation}
    r_{f}(y_i) =
    \begin{cases}
    1, & \text{if } \texttt{prefix}(y_i) = p^*, \\[8pt]
    0.5, & \text{if } \texttt{prefix}(y_i) \neq p^* \\
    & \text{and} \quad \texttt{prefix}(y_i) \in \{p^{\text{fast}}, p^{\text{slow}}\}, \\[8pt]
    0, & \text{otherwise},
    \end{cases}
    \end{equation}
    where $\texttt{prefix}(y_i)$ denotes the prefix extracted from the generated response. This design provides a graded signal that prioritizes mode consistency while still rewarding valid (though mismatched) prefix usage.
\end{itemize}

\paragraph{Optimization objective.}
We optimize the policy model $\pi_\theta$ using the GRPO objective over the $n$ candidates in each group. For each response $y_i$, we compute a relative advantage by subtracting the group mean reward:
\[
A_i = r(y_i) - \operatorname{mean}\big(r(y_1), r(y_2), \dots, r(y_n)\big).
\]
The GRPO objective is computed as:
\begin{equation}
\begin{aligned}
\mathcal{J}_{\text{GRPO}}(\theta)
= \frac{1}{n} \sum_{i=1}^{n} \Bigg[ &
\min \Bigg(
\frac{\pi_{\theta}(y_i\mid x)}{\pi_{\theta_{\text{old}}}(y_i\mid x)} A_i, \;
\operatorname{clip}\!\left(
\frac{\pi_{\theta}(y_i\mid x)}{\pi_{\theta_{\text{old}}}(y_i\mid x)},
1-\epsilon, 1+\epsilon
\right) A_i
\Bigg) \\
& + \beta \mathcal{D}_{\text{KL}}(\pi_{\theta} \parallel \pi_{\text{ref}})
\Bigg],
\end{aligned}
\end{equation}
where $\epsilon$ and $\beta$ are both hyper-parameters. $\epsilon$ controls the tolerance for policy deviation, while $\beta$ determines the strength of the KL penalty, preventing the policy from drifting too far from the reference model $\pi_{\text{ref}}$.

\section{Experiments}

\subsection{Experimental Setup}

\paragraph{Training data.}
We combine multiple public datasets covering 
general visual understanding~\cite{schwenk2022okvqa, lu2022learn, singh2019towards}, spatial reasoning~\cite{lindstrom2022clevr}, chart and document understanding~\cite{kembhavi2016diagram, masry2022chartqa, lu2022dynamic, lu2021iconqa, mathew2021docvqa}, and mathematical reasoning \cite{meng2025mm, wang2025vl}. After applying the dual-mode anchoring process, we end up with a dataset containing 37,506 visual question-answer pairs, among which 18,778 are slow-thinking samples and 18,728 are fast-thinking samples. The detailed composition of the training dataset is provided in the supplementary.

\paragraph{Benchmarks.}
We evaluate our approach on a wide range of multimodal benchmarks. For mathematical reasoning, we choose MathVista \cite{lu2023mathvista} (Testmini) and MathVision \cite{wang2024measuring} (Test). For general visual understanding, we evaluate on MMStar \cite{chen2024we} and MMbench (EN) \cite{liu2024mmbench}. For scientific QA, we use ScienceQA \cite{lu2022learn} and AI2D \cite{kembhavi2016diagram}.

\paragraph{Implementation details.}
We adopt Qwen2.5-VL-7B \cite{bai2025qwen2} as our base model. Training is performed using the TRL~\cite{vonwerra2022trl} framework. For the group reward computation, we sample $n=8$ completions per question. We set the learning rate to $1\times10^{-6}$, rollout batch size to 256, KL coefficient to $1\times10^{-3}$, and maximum generation length to 2,048 tokens.

\subsection{Main Results}

\begin{table*}[t] 
    \centering 
    \caption{\textbf{Comparison of DualMindVLM with state-of-the-art visual reasoning models}. For each benchmark, we report accuracy (acc, \%) and average response length (len, \#tokens). The best result is highlighted in \best{bold}. DualMindVLM strikes the best balance between accuracy and token efficiency among all models.
    }
    \label{tab:main_results} 
    \scriptsize
    \setlength{\tabcolsep}{1.2pt}
    \begin{tabular*}{\textwidth}{l l l *{12}{l}}
    \toprule \multirow{2}{*}{Model} & \multirow{2}{*}{Size} & \multirow{2}{*}{Strategy} & \multicolumn{2}{c}{MathVista} & \multicolumn{2}{c}{MathVision} & \multicolumn{2}{c}{MMStar} & \multicolumn{2}{c}{MMBench} & \multicolumn{2}{c}{ScienceQA} & \multicolumn{2}{c}{AI2D}  \\ 
    \cmidrule(lr){4-5} \cmidrule(lr){6-7} \cmidrule(lr){8-9} \cmidrule(lr){10-11} \cmidrule(lr){12-13} \cmidrule(lr){14-15}& & & {acc $\uparrow$} & {len $\downarrow$} & {acc $\uparrow$} & {len $\downarrow$} & {acc $\uparrow$} & {len $\downarrow$} & {acc $\uparrow$} & {len $\downarrow$} & {acc $\uparrow$} & {len $\downarrow$} & {acc $\uparrow$} & {len $\downarrow$} \\ 
    \midrule 
    \textcolor{gray!50}{Qwen2.5-VL} & \textcolor{gray!50}{7B} & \textcolor{gray!50}{\text{-}} & \textcolor{gray!50}{68.2} & \textcolor{gray!50}{205} & \textcolor{gray!50}{25.1} & \textcolor{gray!50}{511} & \textcolor{gray!50}{63.9} & \textcolor{gray!50}{155} & \textcolor{gray!50}{83.0} & \textcolor{gray!50}{73} & \textcolor{gray!50}{84.0} & \textcolor{gray!50}{156} & \textcolor{gray!50}{80.8} & \textcolor{gray!50}{145}  \\
    LLaVA-CoT & 11B & SFT & 54.8 & 350 & \text{-} & \text{-} & 57.6 & 464 & \text{-} & \text{-} & \text{-} & \text{-} & \text{-} & \text{-}\\
    R1-Onevision & 7B & SFT+RL & 64.1 & 279 & 29.9 & 560 & \text{-} & \text{-} & \text{-} & \text{-} & \text{-} & \text{-} & \text{-} & \text{-} \\
    R1-VL & 7B & SFT+RL & 63.5 & 263 & 24.7 & 363 & 60.0 & 221 & \text{-} & \text{-} & \text{-}& \text{-} & \text{-} & \text{-}  \\
    OpenVLThinker& 7B & SFT+RL & 72.3 & 242 & 25.9 & \best{326} & 63.3 & 200 & 87.5 & 177 & 82.2 & 171 & 83.2 & 160  \\
    MM-Eureka & 7B & RL & 73.0 & 252 & 26.9 & 612 & 64.1 & 246 & 87.3 & 159 & 83.5 & 202 & 83.5 & 207  \\
    ThinkLite & 7B & RL & 75.1 & 247 & 28.5 & 599 & 65.0 & 175 & \best{88.7} & 113 & \text{-} & \text{-} & 83.6 & 168 \\
    VL-Rethinker & 7B & RL & 74.9 & 268 & \best{32.3} & 566 & 64.9 & 231 & 87.6 & 201 & 85.5 & 205 & 82.4 & 226 \\
    \rowcolor{SkyBlue!20}{DualMindVLM} & 7B & RL & \best{75.6} & \best{184} & 30.2 & 446 & \best{65.3} & \best{121} & 88.3 & \best{69} & \best{87.2} & \best{98} & \best{83.8} & \best{104}  \\
    \bottomrule 
    \end{tabular*} 
    \vspace{-3pt} 
\end{table*}

Table \ref{tab:main_results} presents a detailed comparison of our DualMindVLM against state-of-the-art visual reasoning models of similar sizes. Note that all models except LLaVA-CoT and R1-VL are based on the same model, i.e., Qwen2.5-VL. Overall, DualMindVLM achieves state-of-the-art performance while exhibiting exceptionally high token efficiency.

\paragraph{Comparison with the base model.}
Compared with the base model Qwen2.5-VL, DualMindVLM obtains significant improvement in accuracy on \textit{all} benchmarks. Specifically, DualMindVLM improves the accuracy by +7.4\% on MathVista, +5.1\% on MathVision, +1.4\% on MMStar, +5.3\% on MMBench, +3.2\% on ScienceQA, and +3.0\% on AI2D. It is also worth mentioning that DualMindVLM's average output length is shorter than the base model across all benchmarks. These results strongly demonstrate the effectiveness and efficiency of our model.

\paragraph{Comparison with leading reasoning models.}
We compare DualMindVLM with the latest state-of-the-art reasoning models, including VL-Rethinker~\cite{wang2025vl}, ThinkLite~\cite{wang2025sota}, MM-Eureka~\cite{meng2025mm}, OpenVLThinker~\cite{deng2025openvlthinker}, R1-VL~\cite{zhang2025r1}, R1-Onevision~\cite{yang2025r1}, and LLaVA-CoT~\cite{xu2025llava}. In terms of accuracy, DualMindVLM beats the best-performing rivals on four out of six benchmarks, namely MathVista, MMStar, ScienceQA, and AI2D. On MathVision and MMBench, DualMindVLM's performance is close to state-of-the-art. In terms of token usage, DualMindVLM outperforms the reasoning models on all benchmarks except on MathVision where OpenVLThinker produces the least tokens. Compared with the best-performing rival on each benchmark, DualMindVLM reduces token usage by 40\% on average. Overall, DualMindVLM achieves the best balance between accuracy and token efficiency.

\subsection{Ablation Study}

\begin{table}[t]
  \scriptsize
  \small 
  \setlength{\tabcolsep}{4pt}
  \caption{{Ablation study on key components of DualMindVLM.}}
  \label{tab:ablation_component}
  \begin{tabular*}{\linewidth}{@{\extracolsep{\fill}} l c c c c c c c c}
    \toprule
    \multirow{2}{*}{Model} & \multicolumn{2}{c}{MathVista} & \multicolumn{2}{c}{MathVision} & \multicolumn{2}{c}{MMStar} & \multicolumn{2}{c}{ScienceQA}  \\
    \cmidrule(lr){2-3}
    \cmidrule(lr){4-5}
    \cmidrule(lr){6-7}
    \cmidrule(lr){8-9}
     & {acc $\uparrow$} & {len $\downarrow$} & {acc $\uparrow$} & {len $\downarrow$} & {acc $\uparrow$} & {len $\downarrow$} & {acc $\uparrow$} & {len $\downarrow$}  \\
    \midrule
     Qwen2.5-VL-7B & 68.2 & 205 & 25.1 & 511 & 63.9 & 155 & 84.0 & 156 \\
     DualMindVLM-7B & {75.6} & {184} & {30.2} & {446} & {65.3} & {121} & {87.2} & {98} \\
     w/o anchoring & 72.6 & 120 & 28.5 & 332 & 64.9 & 78 & 86.6 & 60 \\
     w/o dual-mode RL & 75.0 & 271 & 28.9 & 584 & 65.2 & 229 & 87.3 & 211  \\
    \bottomrule
  \end{tabular*}
\end{table}

\begin{figure}[t]
\centering

\begin{minipage}{0.48\columnwidth}
    \centering
    \includegraphics[width=\linewidth]{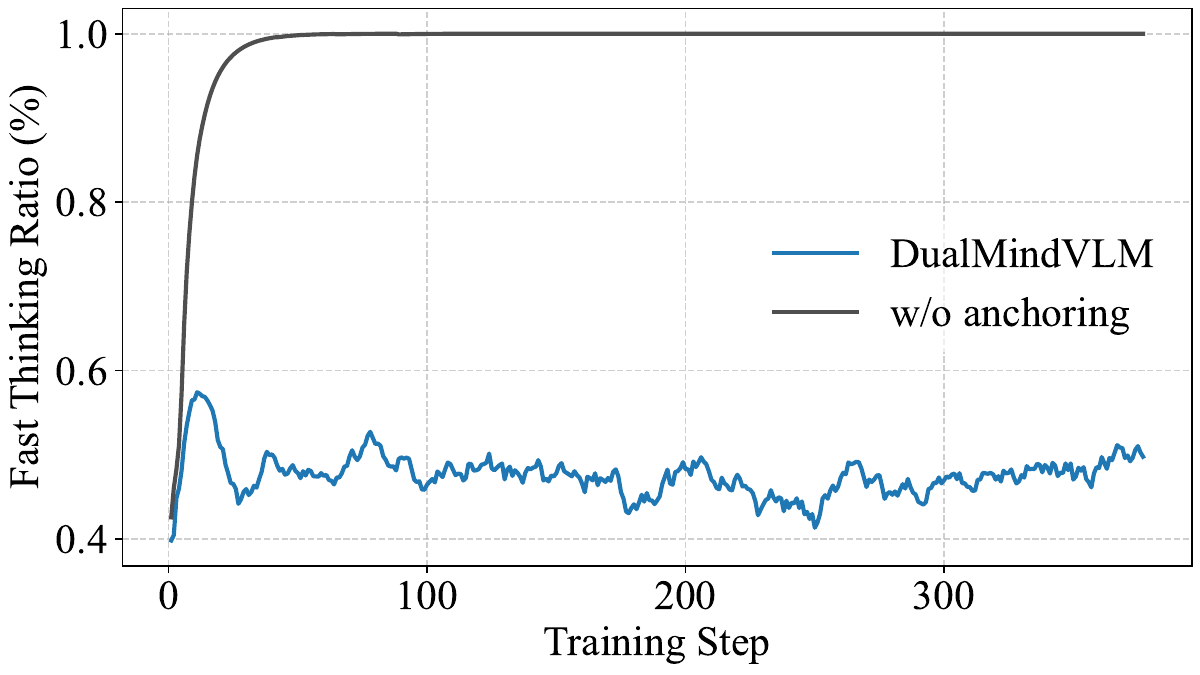}
    \captionof{figure}{\textbf{Fast thinking ratios recorded during training.} Without mode anchoring, the model quickly collapses to the fast thinking mode only, whereas the complete model keeps the ratio well-balanced at around 50\%.}
    \label{fig:exp_short_ratio}
\end{minipage}
\hfill
\begin{minipage}{0.48\columnwidth}
    \centering
    \includegraphics[width=\linewidth]{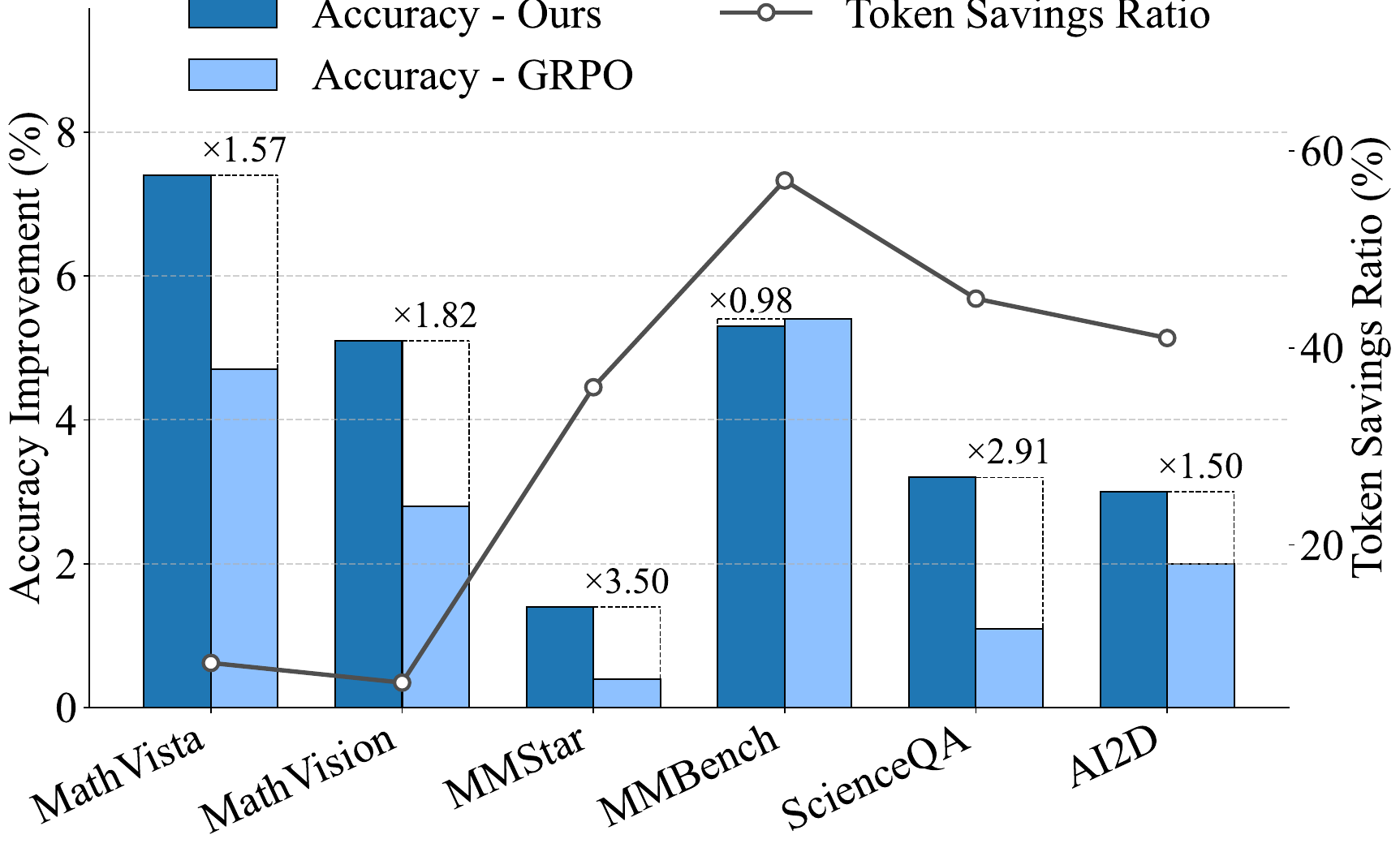}
    \captionof{figure}{\textbf{DualMindVLM vs. GRPO.} We report the performance improvements of DualMindVLM and the GRPO model compared to the base model, along with the token savings ratio relative to GRPO.}
    \label{fig:exp_vs_grpo}
\end{minipage}

\end{figure}

\paragraph{Effect of dual-mode anchoring.}
Recall that our approach consists of two stages: dual-mode anchoring and dual-mode learning (see Fig.~\ref{fig:framework}). We first evaluate the role of dual-mode anchoring. By removing this stage---meaning that we rely only on the dual-mode system prompt shown in Fig.~\ref{fig:sys_prompt} to develop the dual-thinking systems. The results are shown in Table~\ref{tab:ablation_component}. A consistent performance decline is observed across all benchmarks. Most notably, MathVista and MathVision exhibit significant drops of 3.0\% and 1.7\%, respectively. This degradation is primarily attributed to mode collapse during the early stages of RL. Without the guidance of prefix-conditioned rollouts, the model tends to exploit the thinking mode with higher initial likelihood (i.e., the fast-thinking mode, see Fig.~\ref{fig:exp_short_ratio}). This collapse is evidenced by a substantial reduction in average response length, which limits the development of System 2 reasoning and degrades the overall performance.

\paragraph{Effect of dual-mode RL.}
By removing dual-mode RL, the model is trained with GRPO on the same data to just develop System 2 thinking, guided by the prompt \textit{“Please reason step by step”}. As shown in Table~\ref{tab:ablation_component}, accuracy drops on MathVista and MathVision compared with the full model. On MMStar and ScienceQA, performance remains competitive, but response lengths increase substantially, resulting in much lower token efficiency. Notably, even this reduced variant still improves over the base model and surpasses several recent reasoning models shown in Table~\ref{tab:main_results}. This suggests that part of the performance gains may already stem from the mode anchoring stage itself. Since anchoring relies on length-based thresholding, it effectively alters the training data distribution by selecting samples with more pronounced output length tendencies. Data-centric RL for reasoning is beyond the scope of our work. We will investigate this topic in future work.

\paragraph{DualMindVLM vs.~GRPO.}
Fig.~\ref{fig:exp_vs_grpo} compares DualMindVLM with the GRPO model, which is trained without dual-mode anchoring and dual-mode learning. The bar charts show the accuracy improvement over the base model Qwen2.5-VL. while also illustrating that DualMindVLM achieves higher accuracy than GRPO on most benchmarks while using fewer tokens on all benchmarks. The improvements are particularly notable on MMStar and ScienceQA, where the improvement ratios reach 3.50× and 2.91× respectively. On MMBench, DualMindVLM achieves similar accuracy to GRPO, with a slightly lower improvement ratio of 0.99×. However, it provides the largest token savings of around 60\%.

\paragraph{Effect of free-form rollouts.}
As discussed, we use a mixture of free-form and prefix-conditioned rollouts to facilitate the learning of automatic System 1+2 thinking. Table~\ref{tab:free_form_ratio} shows the results obtained by varying the number of free-form generations $m$ during GRPO sampling. We consider three settings: no free-form generation ($m = 0$), half free-form generation ($m = 4$), and full free-form generation ($m = 8$). When no free-form generation is adopted, the model is only guided by a pre-defined thinking mode prefix and therefore struggles to learn how to automate the prefix selection. The model using full free-form generation is equivalent to the model trained without the anchored thinking mode.

\begin{table}[t]
  \centering
  \small
  \setlength{\tabcolsep}{4pt}
  \caption{{Effect of free-form rollouts during GRPO sampling.}}
  \label{tab:free_form_ratio}
  \begin{tabular*}{\linewidth}{@{\extracolsep{\fill}} c c c c c c c c c}
    \toprule
    \multirow{2}{*}{Free-Form Rollouts}
    & \multicolumn{2}{c}{MathVista}
    & \multicolumn{2}{c}{MathVision}
    & \multicolumn{2}{c}{MMStar}
    & \multicolumn{2}{c}{ScienceQA} \\
    \cmidrule(lr){2-3}\cmidrule(lr){4-5}\cmidrule(lr){6-7}\cmidrule(lr){8-9}
     & {acc $\uparrow$} & {len $\downarrow$} & {acc $\uparrow$} & {len $\downarrow$} & {acc $\uparrow$} & {len $\downarrow$} & {acc $\uparrow$} & {len $\downarrow$} \\
    \midrule
     Zero ($m=0$) & 73.4 & 172 & 29.6 & 519 & 64.0 & 145 & 84.5 & 97 \\
     Half ($m=4$)  & {75.6} & 184 & {30.2} & 449 & 65.3 & 121 & 87.2 & 98 \\
     Full ($m=8$)  & 72.6 & 120 & 28.5 & 332 & 64.9 & 78 & 86.6 & 60 \\
    \bottomrule
  \end{tabular*}
\end{table}

\subsection{Further Analysis}
\label{sec:futher_analysis}

\begin{table}[t]
  \centering
  \small
  \setlength{\tabcolsep}{8pt}
  \caption{\textbf{Sensitivity to length thresholds.} Comparison of performance and response length averaged over six benchmarks. ``Fast'' and ``Slow'' denote the average response length in fast and slow thinking, while ``Mean'' denotes the overall mean response length.}
  \label{tab:fast_slow_threshold}
  \begin{tabular}{c c c c c c}
    \toprule
    \multirow{2}{*}{$\tau_{\text{fast}}$} &
    \multirow{2}{*}{$\tau_{\text{slow}}$} & 
    \multirow{2}{*}{Accuracy}
    & \multicolumn{3}{c}{Length}
    \\
    \cmidrule(lr){4-6}
     & & & Fast & Slow & Mean  \\
    \midrule
     None & None & 68.9 & 72 & 131 & 98 \\
     100 & 200 & 70.1 & 23 & 275 & 164  \\
     50 & 200 & 69.8 & 23 & 302 & 197  \\
     100 & 250 & 69.8 & 29 & 290 & 173  \\
    \bottomrule
  \end{tabular}
\end{table}

\paragraph{Threshold sensitivity.} To investigate how the length thresholds ($\tau_{\text{fast}}$ and $\tau_{\text{slow}}$) influence the model's behavior, we sample 5k fast-thinking and 5k slow-thinking examples from the training set according to different thresholds, and train a separate model for each configuration. The ``None'' configuration serves as a baseline, where fast- and slow-thinking samples are randomly selected. 

As shown in Table~\ref{tab:fast_slow_threshold}, two key phenomena emerge. First, randomly assigning thinking modes without considering length-based priors significantly blurs the boundaries between the two thinking modes: the fast-thinking mode becomes notably longer (72 vs. 23 tokens) while the slow-thinking mode shrinks (131 vs. 275 tokens), resulting in a 1.2\% drop in accuracy. Second, our length-based anchoring is robust to the specific choice of thresholds: varying $\tau_{\text{fast}}$ and $\tau_{\text{slow}}$ across rows 2--4 results in only marginal performance fluctuations ($\le 0.3\%$). This indicates that anchoring thinking modes with length-based priors helps maintain a clear separation between the two behaviors.

\begin{figure*}[t]
    \centering
    \begin{subfigure}[t]{0.49\textwidth}
        \centering
        \includegraphics[width=\textwidth]{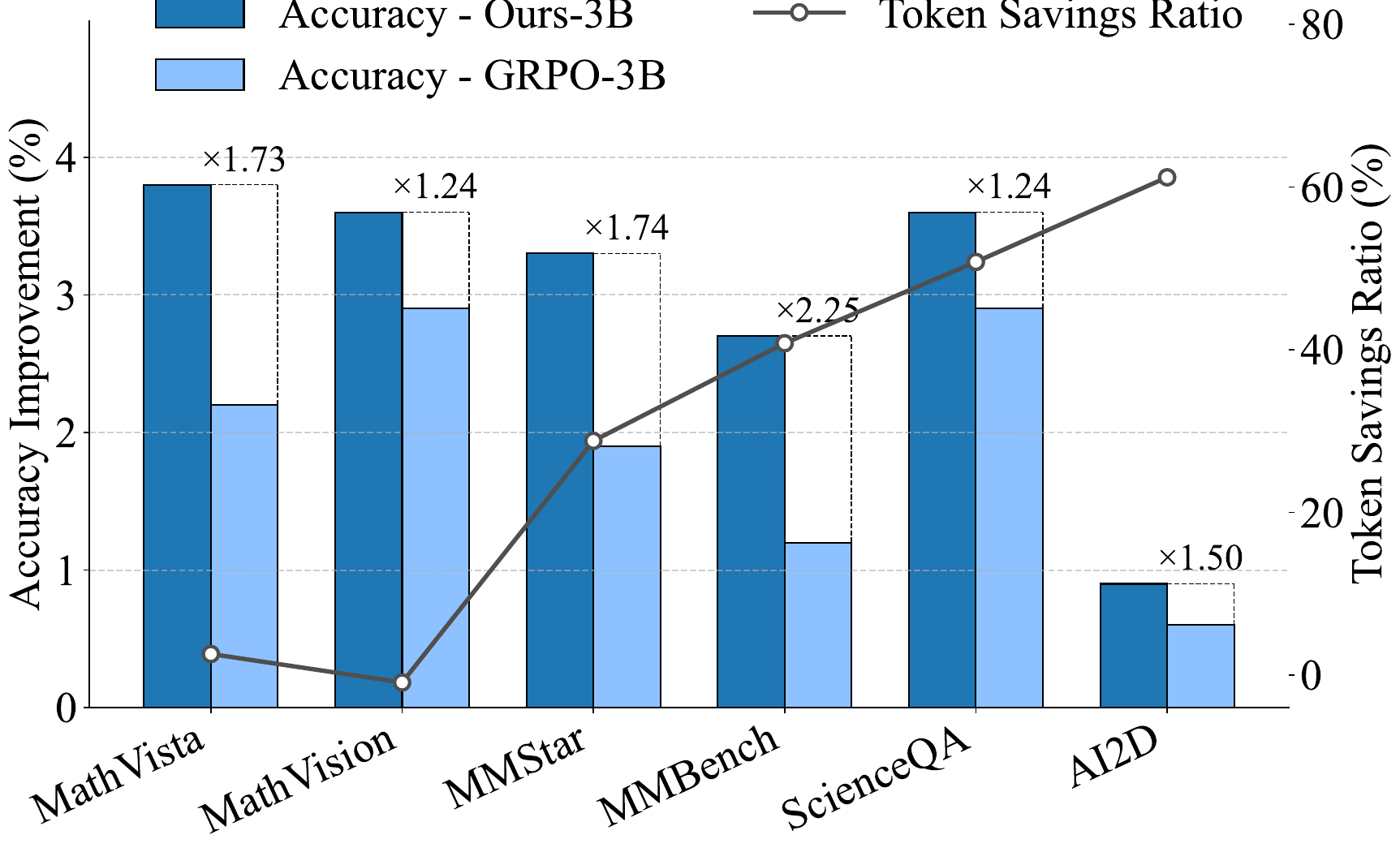}
        \caption{Qwen2.5-VL-3B}
        \label{fig:exp_vs_grpo_3b_sub}
    \end{subfigure}
    \hfill
    \begin{subfigure}[t]{0.49\textwidth}
        \centering
        \includegraphics[width=\textwidth]{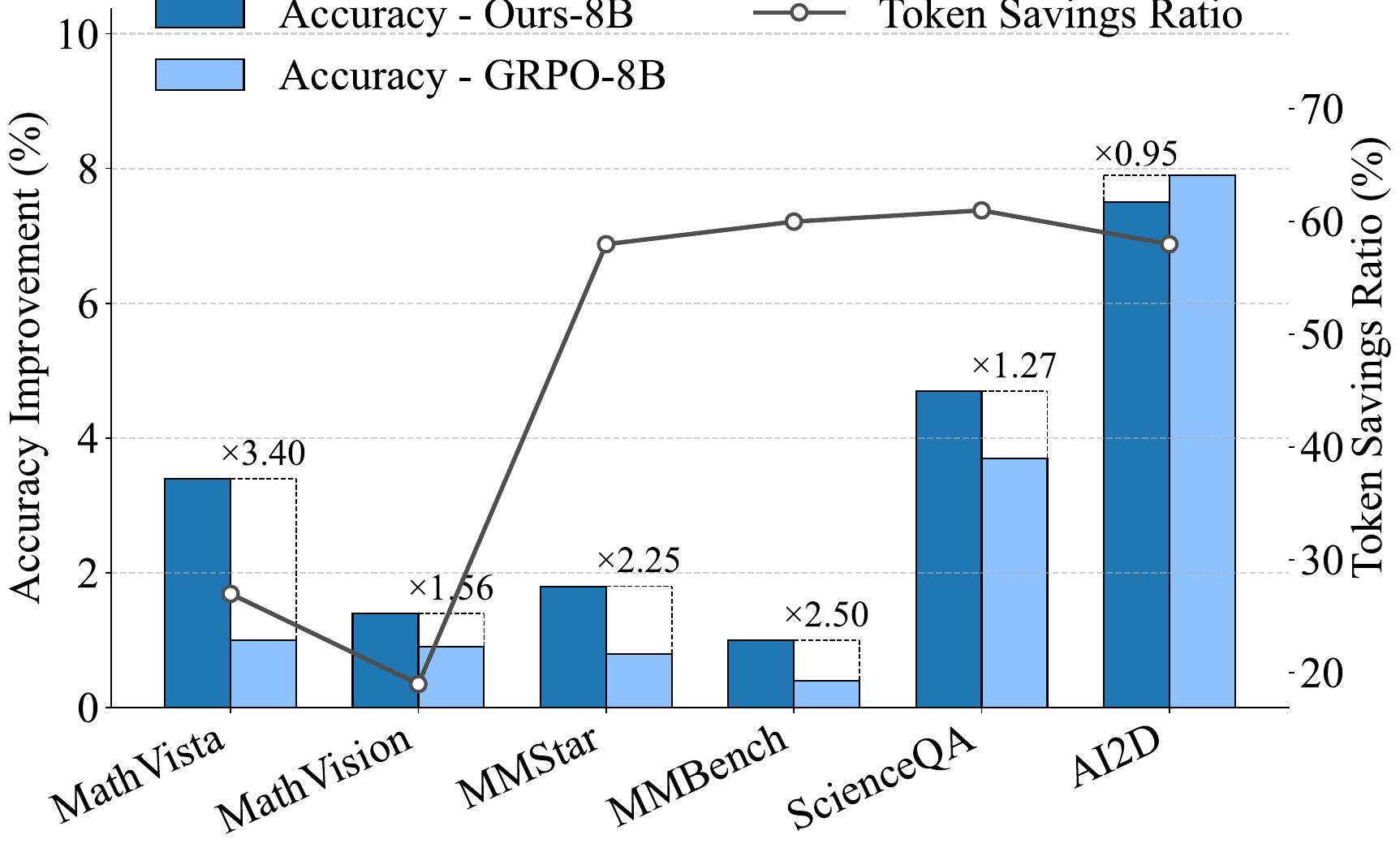}
        \caption{InternVL3-8B}
        \label{fig:exp_vs_grpo_intern_sub}
    \end{subfigure}
    \caption{\textbf{Generalization on architecture and scale.} DualMindVLM consistently outperforms GRPO across most benchmarks while significantly reducing token usage on both Qwen2.5-VL-3B and InternVL3-8B.}
    \label{fig:exp_generalization}
\end{figure*}

\paragraph{Generalization.} To evaluate generalization across architectures and scales, we evaluate our method on two additional backbones: Qwen2.5-VL-3B and InternVL3-8B. As shown in Fig.~\ref{fig:exp_generalization} (Qwen2.5-VL-3B in (a), InternVL3-8B in (b)), DualMindVLM consistently outperforms GRPO on most benchmarks while reducing token usage by up to about 60\%. Notably, the largest improvements are observed on MathVista and MMBench for both backbones: on Qwen2.5-VL-3B the gains reach $1.73\times$ and $2.25\times$, while on InternVL3-8B the improvements further increase to $3.40\times$ and $2.50\times$, respectively.

\begin{table*}[t]
    \centering
    \caption{\textbf{Performance under different thinking modes.} 
Forced-Fast and Forced-Slow enforce the model to always use fast or slow thinking, respectively, while Auto allows the model to select the thinking mode automatically.}
    \label{tab:mode_specific_performance}
    \small
    \setlength{\tabcolsep}{2.1pt}
    \begin{tabular*}{\textwidth}{l *{12}{l}}
    \toprule
    \multirow{2}{*}{Mode} & \multicolumn{2}{c}{MathVista} & \multicolumn{2}{c}{MathVision} & \multicolumn{2}{c}{MMStar} & \multicolumn{2}{c}{MMBench} & \multicolumn{2}{c}{ScienceQA} & \multicolumn{2}{c}{AI2D} \\
    \cmidrule(lr){2-3} \cmidrule(lr){4-5} \cmidrule(lr){6-7} \cmidrule(lr){8-9} \cmidrule(lr){10-11} \cmidrule(lr){12-13}
    & {acc $\uparrow$} & {len $\downarrow$} & {acc $\uparrow$} & {len $\downarrow$} & {acc $\uparrow$} & {len $\downarrow$} & {acc $\uparrow$} & {len $\downarrow$} & {acc $\uparrow$} & {len $\downarrow$} & {acc $\uparrow$} & {len $\downarrow$} \\
    \midrule
    Auto & 75.6 & 184 & 30.2 & 446 & 65.3 & 121 & 88.3 & 69 & 87.2 & 98 & 83.8 & 104 \\
    Forced-Fast & 71.8 & 76 & 27.7 & 271 & 64.4 & 59 & 87.9 & 37 & 86.1 & 51 & 83.1 & 47 \\
    Forced-Slow & 75.7 & 221 & 30.2 & 461 & 65.9 & 160 & 88.6 & 112 & 87.5 & 141 & 84.1 & 135 \\
    \bottomrule
    \end{tabular*}
\end{table*}

\paragraph{Mode behavior.} To better understand the reasoning behaviors in DualMindVLM, we evaluate the model under three inference modes: Auto, Forced-Fast, and Forced-Slow. As shown in Table~\ref{tab:mode_specific_performance}, enforcing slow thinking consistently improves accuracy across all six benchmarks compared with forced fast thinking. This indicates that slow thinking represents a distinct reasoning behavior rather than simply producing longer responses. Meanwhile, the Auto mode achieves accuracy close to Forced-Slow across all benchmarks. In terms of response length, the gap between Auto and Forced-Slow is relatively small on the math benchmarks, while larger reductions are observed on the more general benchmarks. Overall, this indicates that the routing behavior learned from the length-prior-based training achieves a favorable balance between accuracy and efficiency.




\begin{figure*}[t]
    \centering
    \begin{subfigure}[t]{0.49\textwidth}
        \centering
        \includegraphics[width=\textwidth]{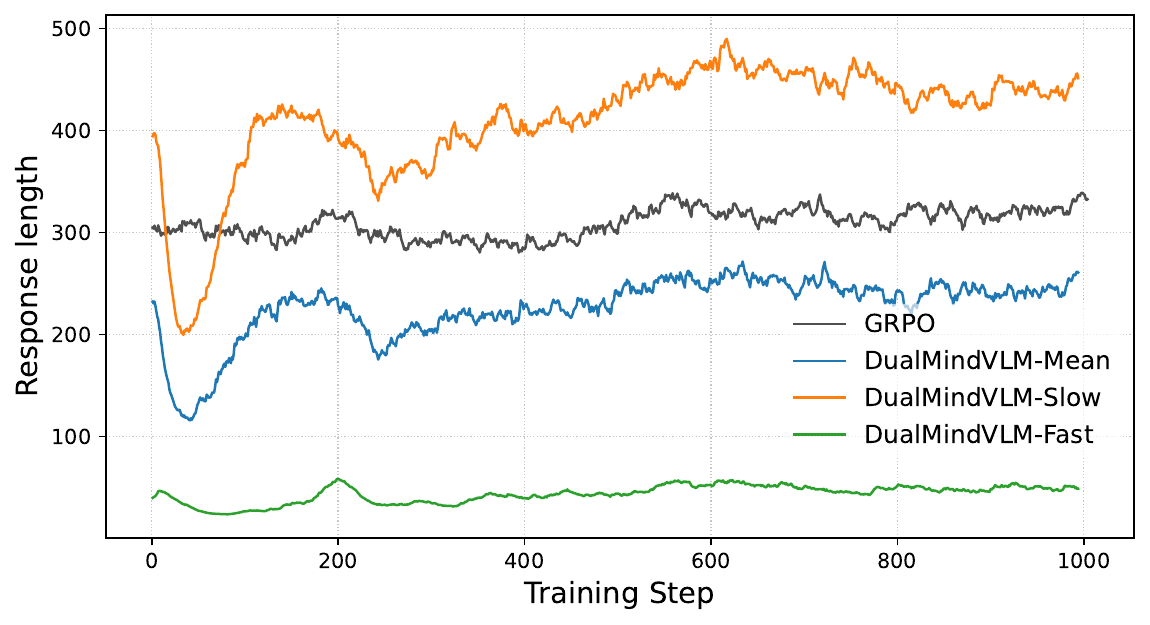}
        \caption{Average response length over training steps}
        \label{fig:sup_length_comparison}
    \end{subfigure}
    \hfill
    \begin{subfigure}[t]{0.49\textwidth}
        \centering
        \includegraphics[width=\textwidth]{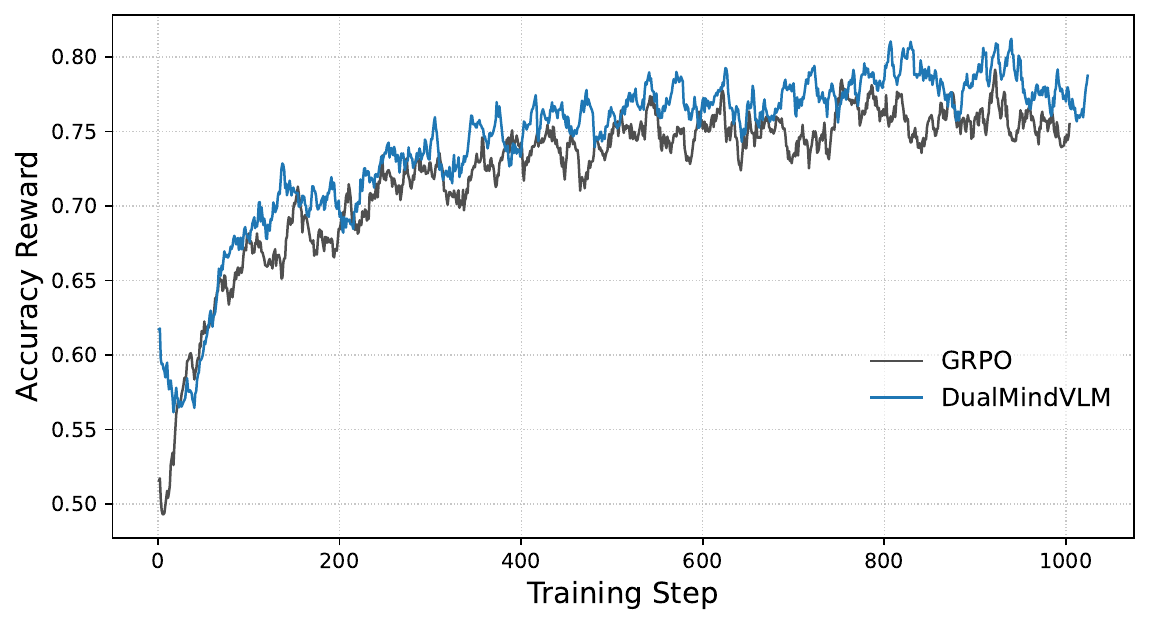}
        \caption{Accuracy reward over training steps}
        \label{fig:sup_acc_comparison}
    \end{subfigure}
    \caption{Training dynamics of response length and accuracy reward during RL training}
    \label{fig:sup_training_dynamics}
\end{figure*}

\paragraph{Training dynamics.}
We report the training dynamics of DualMindVLM and GRPO during RL optimization. 
Fig.~\ref{fig:sup_training_dynamics}~(a) shows the response length dynamics over training steps. DualMindVLM-Fast and DualMindVLM-Slow denote the average response lengths of the fast and slow thinking modes, while DualMindVLM-Mean represents the overall average length. With mode anchoring and the hybrid group sampling strategy, DualMindVLM explores two thinking modes with clearly separated response lengths. The initial drop followed by a rapid recovery suggests that the model quickly adapts to the prefix-conditioned training signals in the early stage of RL. Moreover, the average lengths of both thinking modes remain consistently shorter than those of GRPO throughout training, indicating a potential efficiency advantage of dual-mode optimization.
Fig.~\ref{fig:sup_training_dynamics}~(b) presents the accuracy reward dynamics during training. DualMindVLM consistently achieves higher accuracy reward than GRPO, suggesting that exploring thinking modes aligned with the model's inherent length tendency can lead to more effective optimization and partially explain the performance advantage of DualMindVLM.

\begin{table}[t]
  \centering
  \small
  \setlength{\tabcolsep}{4.3pt}
  \caption{Comparison of visual reasoning models on HumbleBench. DualMindVLM performs the best, meaning that dual-mode thinking has potential to mitigate hallucinations.}
  \label{tab:humblebench}
  \begin{tabular}{l c c c c}
  \toprule
  \multirow{2}{*}{Model} & \multicolumn{4}{c}{HumbleBench} \\
  \cmidrule(lr){2-5}
  & Relation $\uparrow$ & Attribute $\uparrow$ & Object $\uparrow$ & Overall $\uparrow$ \\
  \midrule
  R1-Onevision  & 65.2 & 73.4 & 61.4 & 66.9 \\
  R1-VL  & 68.0 & 74.0 & 63.6 & 68.7  \\
  MM-Eureka & 63.2 & 74.7 & 64.1 & 67.5 \\
  ThinkLite & 69.5 & 77.2 & 66.8 & 71.3 \\
  VL-Rethinker  & 68.3 & 76.6 & 65.2 & 70.3  \\
  \rowcolor{SkyBlue!20}DualMindVLM  & \best{70.0} & \best{77.5} & \best{67.0} & \best{71.7} \\
  \bottomrule
  \end{tabular}
\end{table}

\paragraph{Hallucination.}
Longer reasoning chains are known to have a higher risk of producing hallucinated answers~\cite{liu2025more}. We evaluate DualMindVLM as well as five other reasoning VLMs on HumbleBench~\cite{tong2025measuring}, a hallucination benchmark consisting of 22,831 multiple-choice questions and covering hallucinations in relation, attribute, and object. Notably, each question includes a ``None of the above'' option, requiring the model to not only recognize correct visual information but also refuse to choose when all answers are incorrect. Table~\ref{tab:humblebench} shows that DualMindVLM beats all the competitors by a clear margin across all hallucination types. These results strongly demonstrate that developing dual thinking modes, rather than uniformly encouraging long reasoning, is more effective in mitigating hallucinations.

\paragraph{Limitations.} While DualMindVLM improves the balance between reasoning performance and inference efficiency, our framework primarily structures reasoning within the language space. As a result, the model relies on language-mediated reasoning rather than directly re-examining visual evidence during inference. This limitation may affect tasks that require fine-grained perceptual grounding or detailed visual verification. A promising future direction is to integrate our dual-mode framework with ``think-with-image'' mechanisms, enabling models to dynamically revisit visual features during the reasoning process while maintaining high computational efficiency.

\section{Conclusion}

In this paper, we propose DualMindVLM, a System~1+2 thinking VLM that develops dual thinking modes derived from the model's inherent length prior. By anchoring and reinforcing distinct reasoning patterns through RL, our approach eliminates the need for external supervision. Experimental results on six challenging multimodal reasoning benchmarks show that DualMindVLM achieves performance on par with state-of-the-art visual reasoning models while using substantially fewer tokens on average. We hope this work encourages future research on developing more efficient and controllable reasoning paradigms for vision-language models.

%
%
\bibliographystyle{splncs04}
\clearpage
\appendix
\makeatletter
\renewcommand*{\theHsection}{appendix.\thesection}
\makeatother

\section{Training Dataset}

\begin{figure}[!htbp]
    \centering
    \includegraphics[width=0.55\columnwidth]{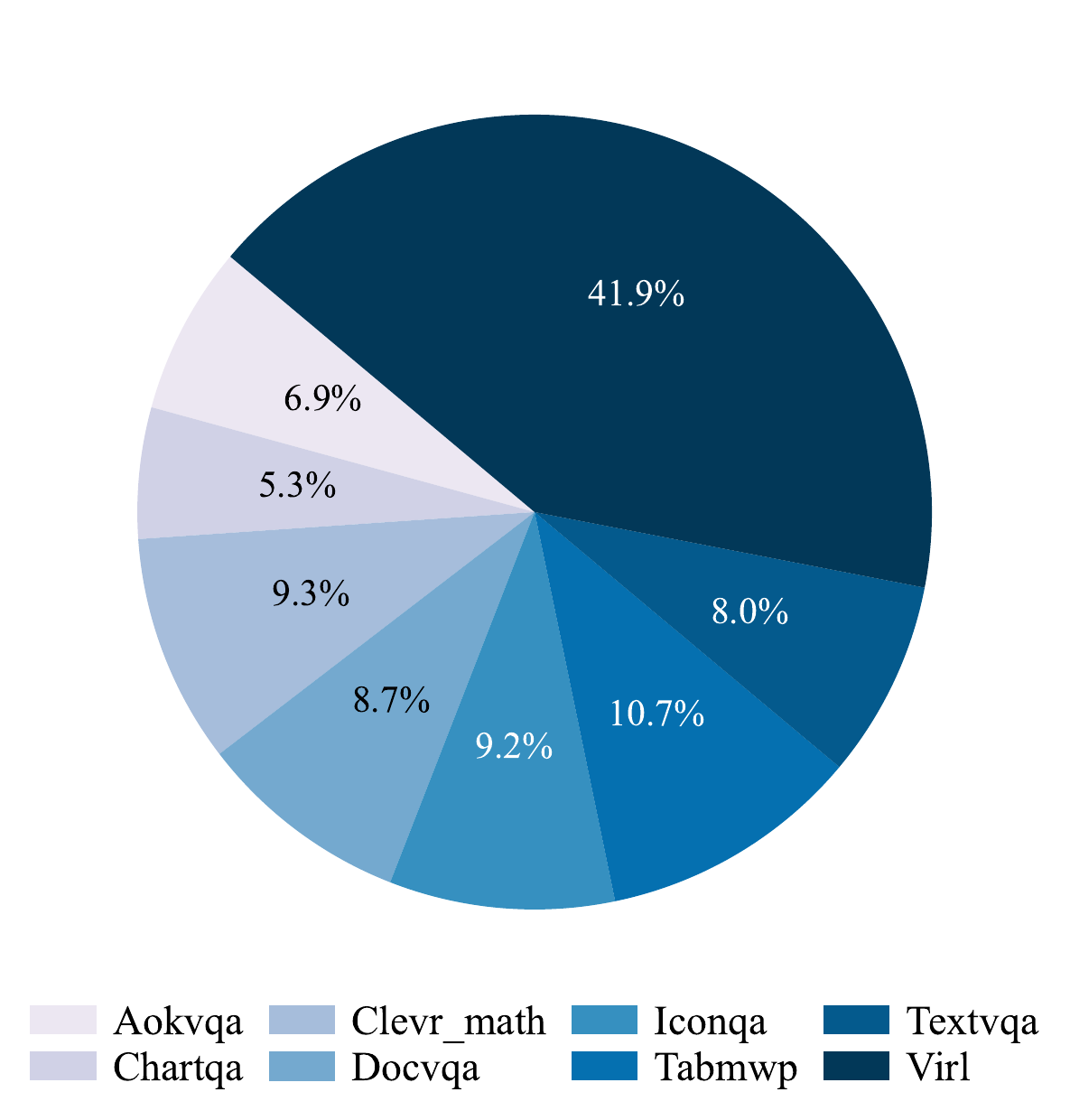}
    \caption{Distribution of training samples across different datasets.}
    \label{fig:sup_dataset_distribution}
\end{figure}

Our training set is constructed by aggregating samples from a diverse collection of visual question answering datasets covering multiple visual reasoning scenarios. As illustrated in Figure~\ref{fig:sup_dataset_distribution}, the data sources include natural scene understanding (A-OKVQA~\cite{schwenk2022okvqa} and TextVQA~\cite{singh2019towards}), structured visual reasoning such as charts and diagrams (ChartQA~\cite{masry2022chartqa} and IconQA~\cite{lu2021iconqa}), document and table understanding (DocVQA~\cite{mathew2021docvqa} and TabMWP~\cite{lu2022learn}), visual counting tasks (CLEVR-Math~\cite{lindstrom2022clevr}), as well as math-intensive reasoning problems (VIRL~\cite{wang2025vl}). This diverse composition allows the training corpus to span a broad spectrum of visual reasoning complexity, ranging from perception-oriented queries to structured and multi-step reasoning problems.

\begin{figure}[t]
    \centering
    \begin{subfigure}[t]{0.48\columnwidth}
        \centering
        \includegraphics[width=\columnwidth]{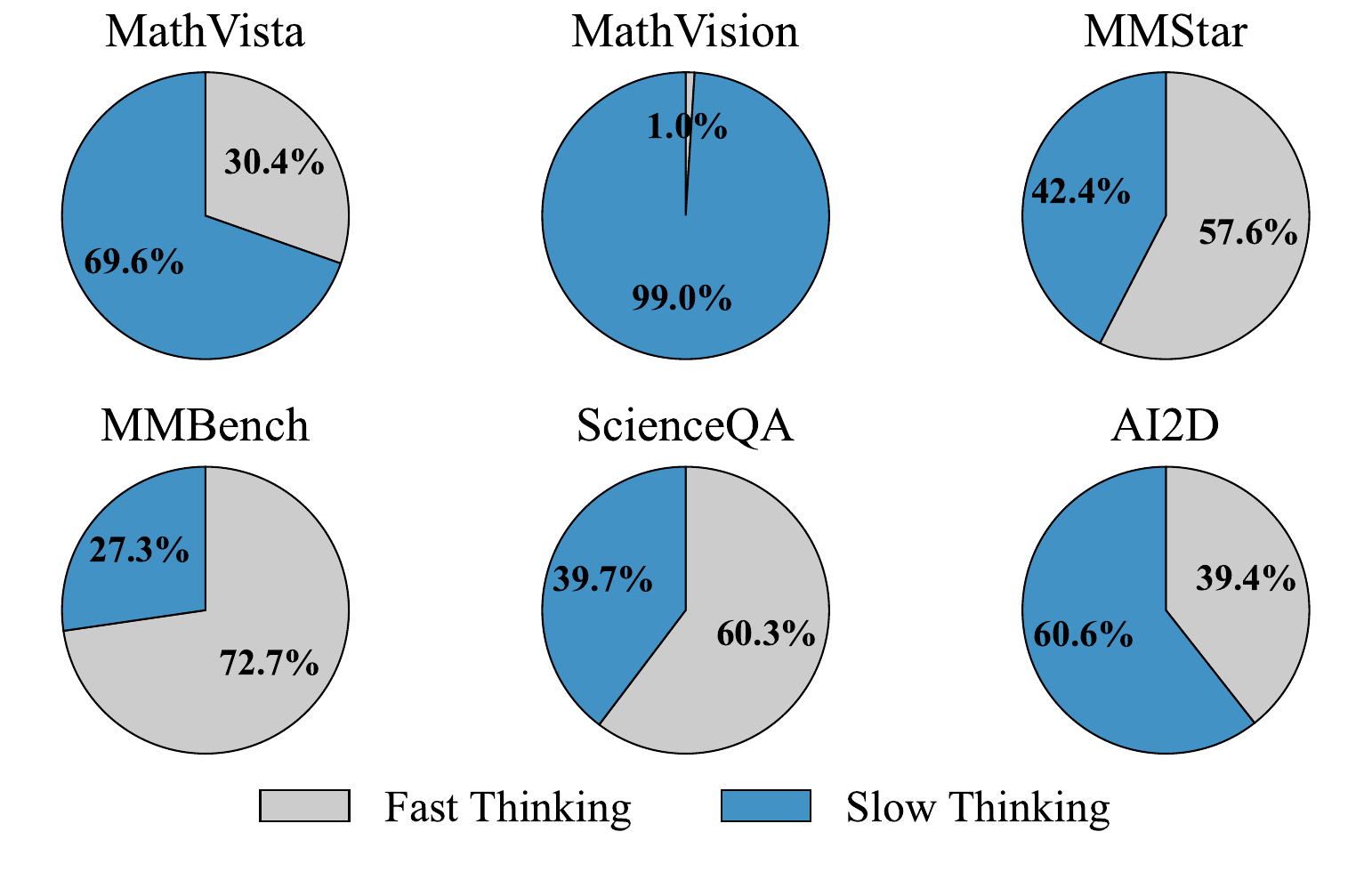}
        \caption{Thinking mode selection ratios.}
        \label{fig:exp_mode_selection_ratios}
    \end{subfigure}
    \hfill
    \begin{subfigure}[t]{0.48\columnwidth}
        \centering
        \includegraphics[width=\columnwidth]{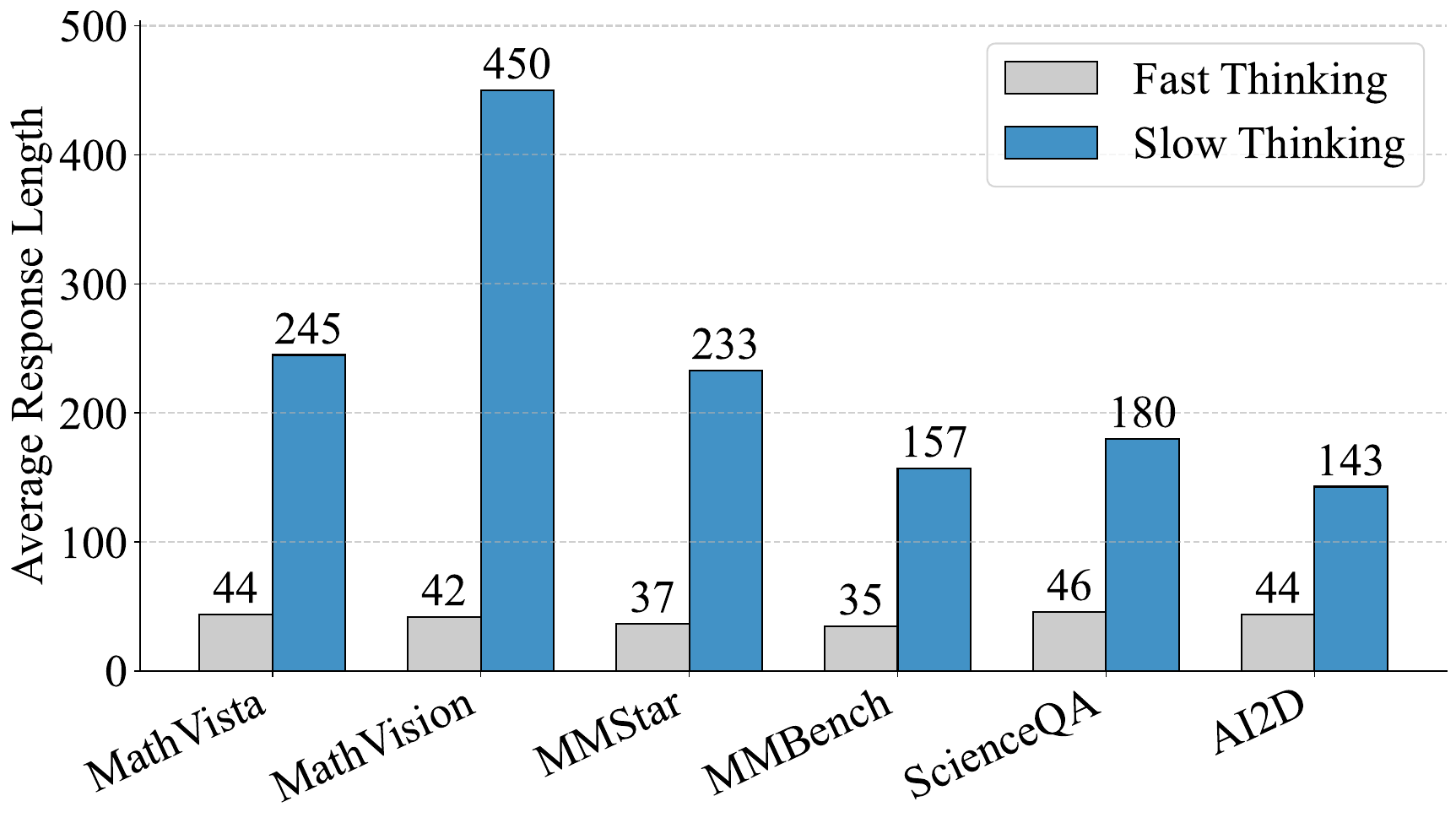}
        \caption{Average response lengths on fast and slow thinking modes.}
        \label{fig:exp_mode_length}
    \end{subfigure}
    \caption{\textbf{Thinking mode selection of DualMindVLM across different tasks.} (a) The model adaptively selects fast or slow thinking depending on task characteristics. (b) Fast-thinking responses remain concise, while slow-thinking responses exhibit length variation across tasks.}
    \label{fig:exp_mode_analysis}
\end{figure}

\begin{figure}[t]
    \centering
    \includegraphics[width=0.65\columnwidth]{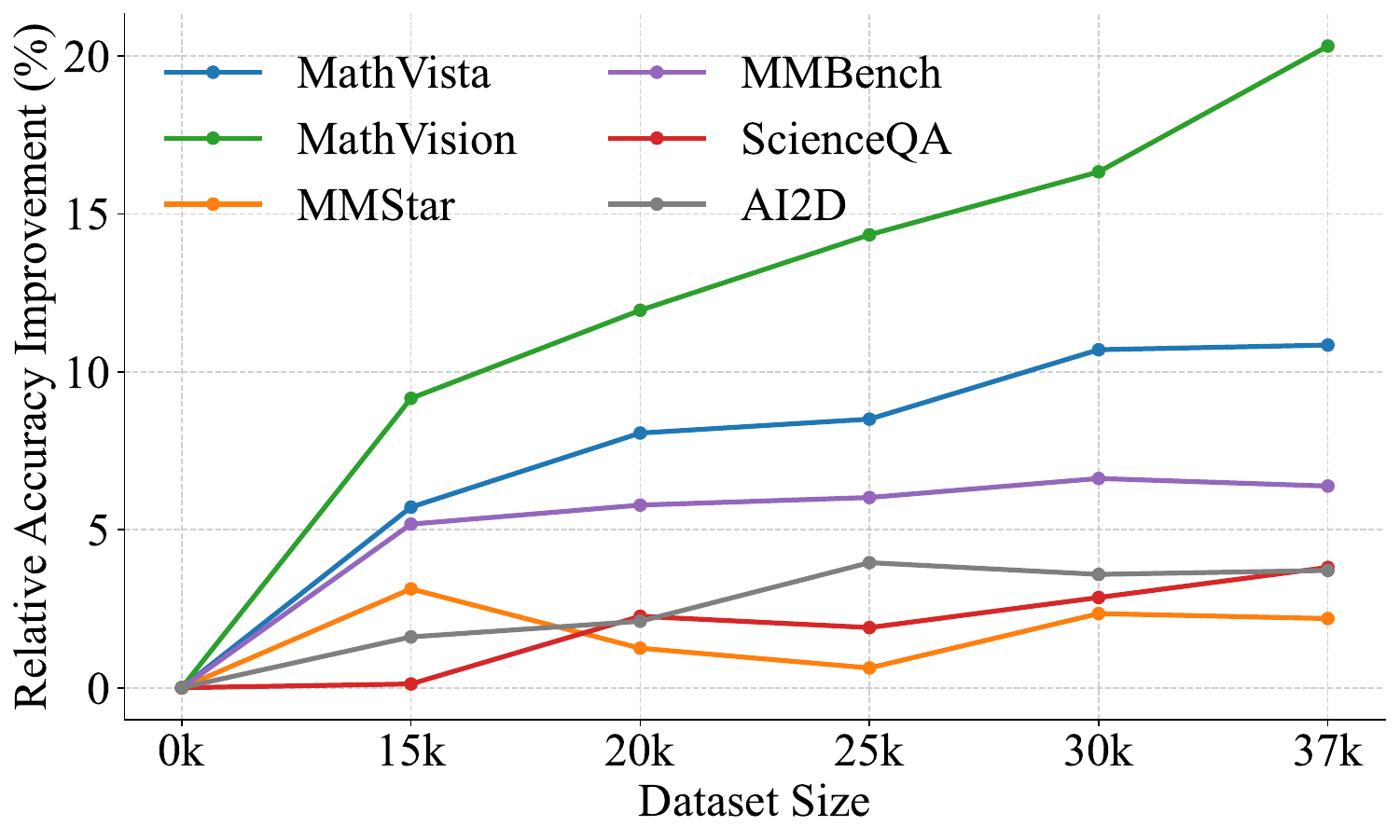}
    \caption{\textbf{Effect of training dataset scale.} Larger scale benefits complex problems like math. The impact is limited for simpler problems.}
    \label{fig:datasize}
\end{figure}

\section{Thinking Mode Selection}

We calculate the ratios between fast and slow thinking modes automatically selected by DualMindVLM during inference, as well as the average output lengths for both modes. The thinking mode selection ratios are presented in Figure~\ref{fig:exp_mode_selection_ratios}. As expected, the model favors the slow thinking mode for challenging problems like math (MathVista and MathVision) and exhibits a relatively balanced mode selection behavior on other benchmarks. Figure~\ref{fig:exp_mode_length} reports the average output lengths on the six benchmarks. In general, the output generated in fast thinking mode remains below 50 tokens, demonstrating stable and concise thinking behavior. In contrast, the slow thinking mode leads to responses of varying lengths that reflect different thinking efforts for different types of problems.

\section{Effect of Dataset Scale}

We explore how the training dataset scale affects the performance during our dual-mode training. To this end, we vary the number of samples used for training DualMindVLM. Specifically, we start from 15k and then gradually increase the number to 37k. In all settings, the training data maintain a balanced composition of fast-thinking and slow-thinking samples. The results are shown in Figure~\ref{fig:datasize} where the accuracy improvement is calculated relative to the base model. We have made some intriguing observations. Increasing the scale does not always yield better results. Specifically, for challenging problems like those in MathVista and MathVision, expanding the dataset proves beneficial, as evidenced by the clear upward trends in both curves. In contrast, for scientific or perceptual tasks such as ScienceQA, AI2D, MMBench, and MMStar, performance gains with increasing data are limited or fluctuate.

\section{Case Study}

\begin{figure*}[htbp]
    \centering
    \includegraphics[width=\textwidth]{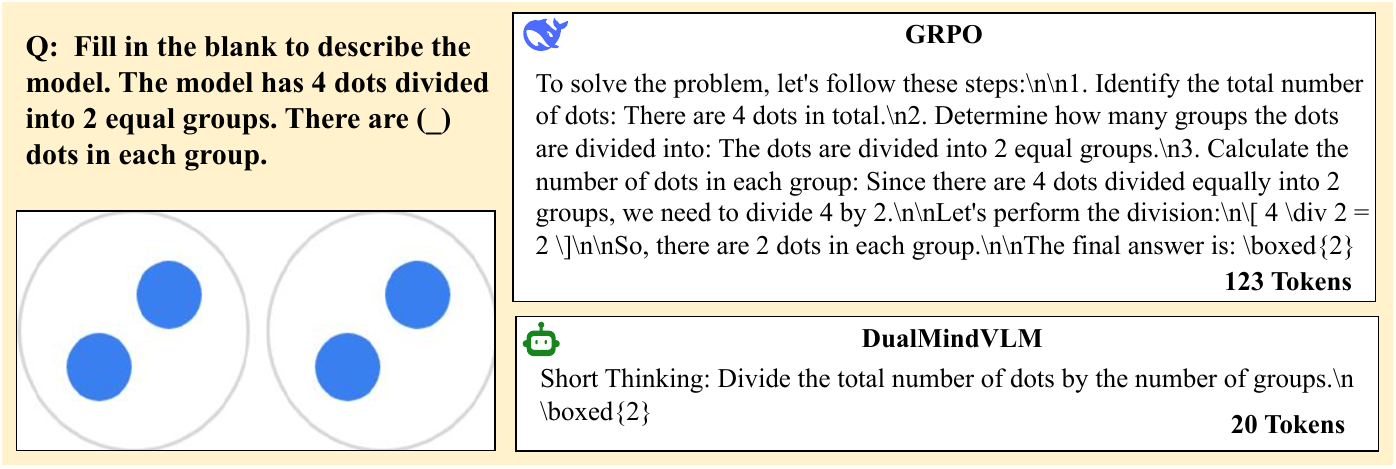}
    \caption{Example responses of the GRPO model and DualMindVLM to a diagram-based VQA question.}
    \label{fig:sup_case1}
\end{figure*}

\begin{figure*}[htbp]
    \centering
    \includegraphics[width=\textwidth]{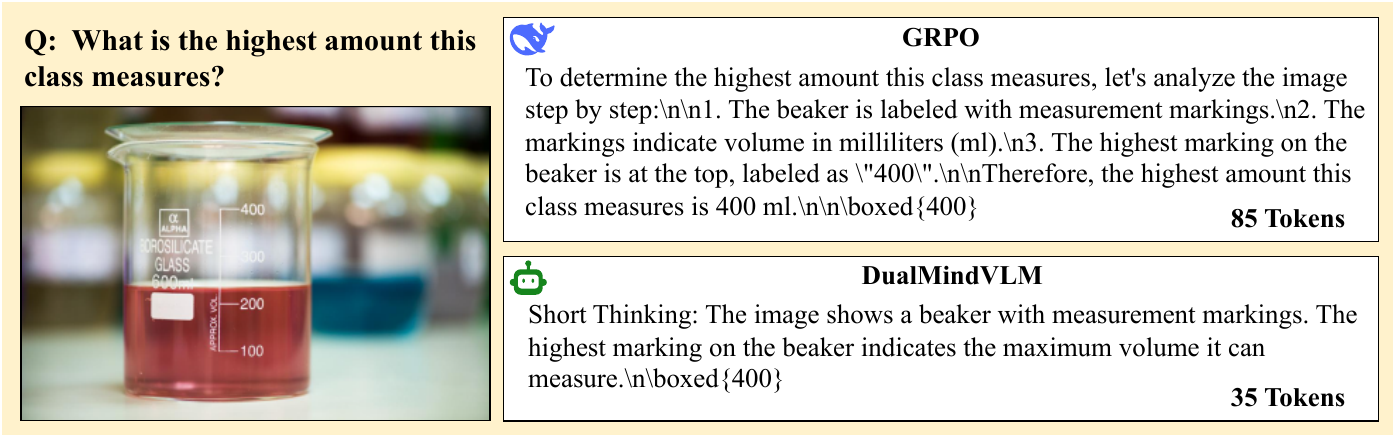}
    \caption{Example responses of the GRPO model and DualMindVLM to a general scene-based VQA question.}
    \label{fig:sup_case2}
\end{figure*}

\begin{figure*}[htbp]
    \centering
    \includegraphics[width=\textwidth]{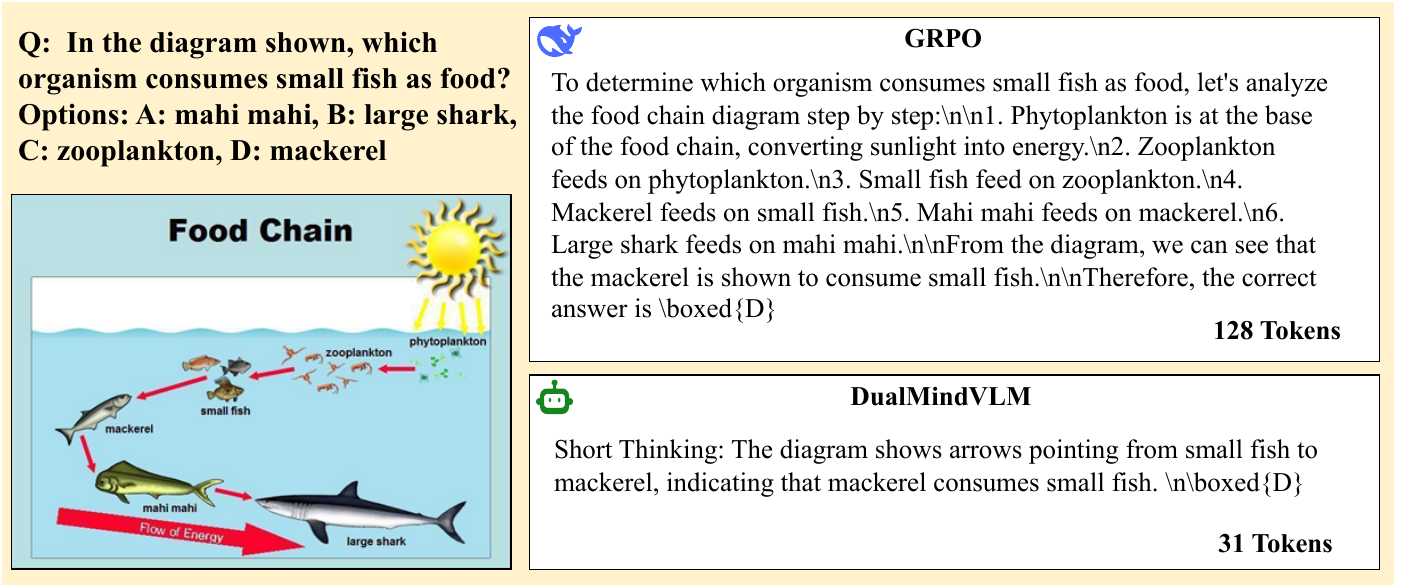}
    \caption{Example responses of the GRPO model and DualMindVLM to a scientific VQA question.}
    \label{fig:sup_case3}
\end{figure*}

\begin{figure*}[htbp]
    \centering
    \includegraphics[width=\textwidth]{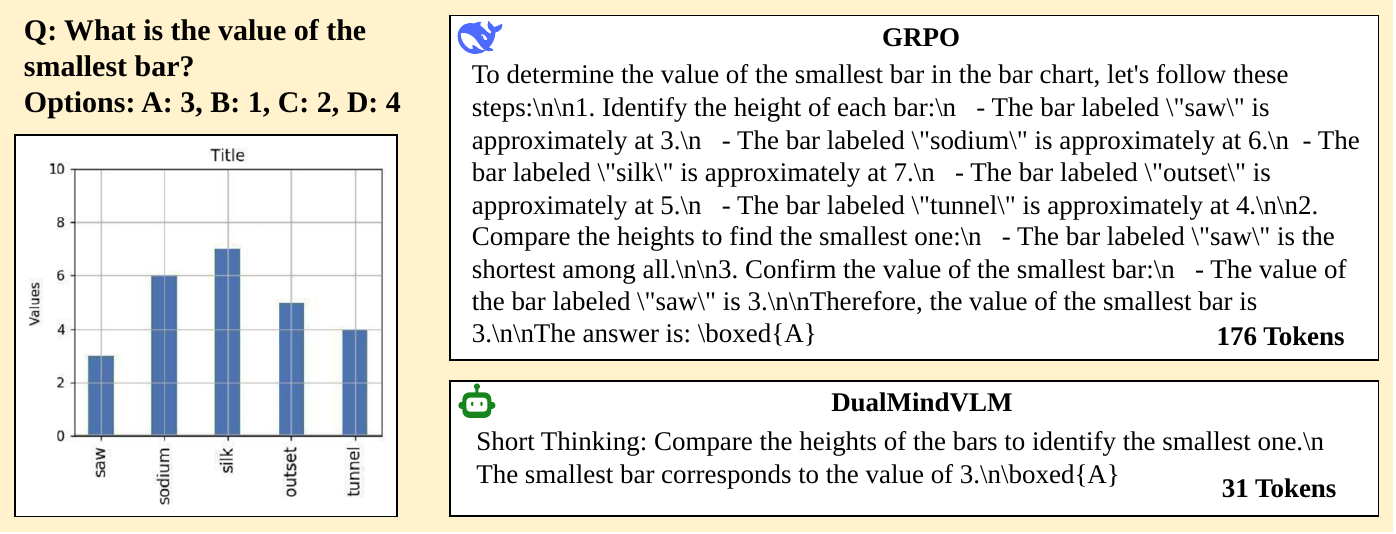}
    \caption{Example responses of the GRPO model and DualMindVLM to a chart-based VQA question.}
    \label{fig:sup_case4}
\end{figure*}

\begin{figure*}[htbp]
    \centering
    \includegraphics[width=\textwidth]{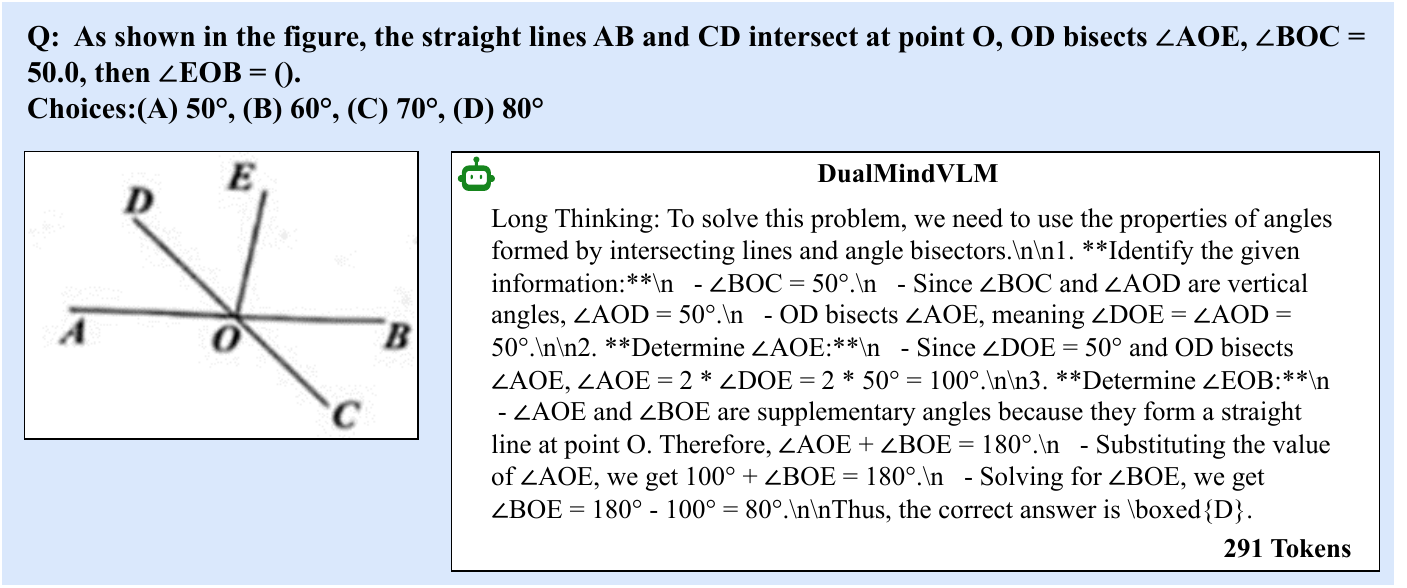}
    \caption{An example response of DualMindVLM to a geometric reasoning VQA question.}
    \label{fig:sup_case5}
\end{figure*}

\begin{figure*}[htbp]
    \centering
    \includegraphics[width=\textwidth]{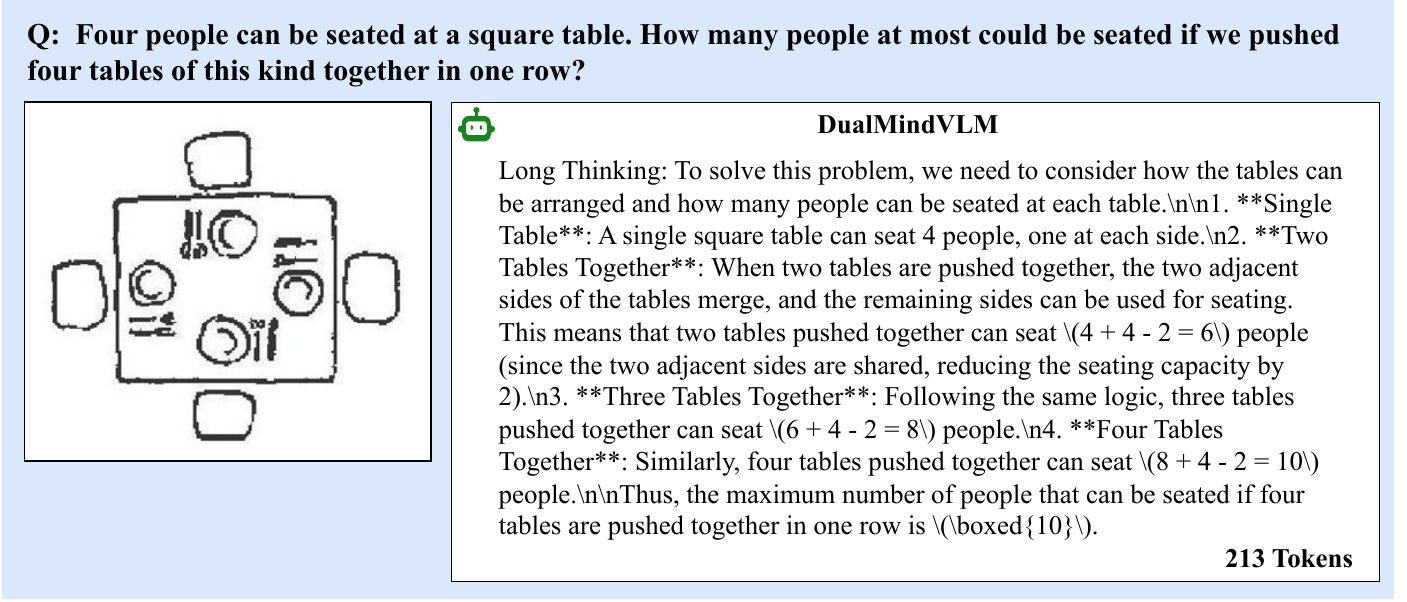}
    \caption{An example response of DualMindVLM to a logic reasoning VQA question.}
    \label{fig:sup_case6}
\end{figure*}

\begin{figure*}[htbp]
    \centering
    \includegraphics[width=\textwidth]{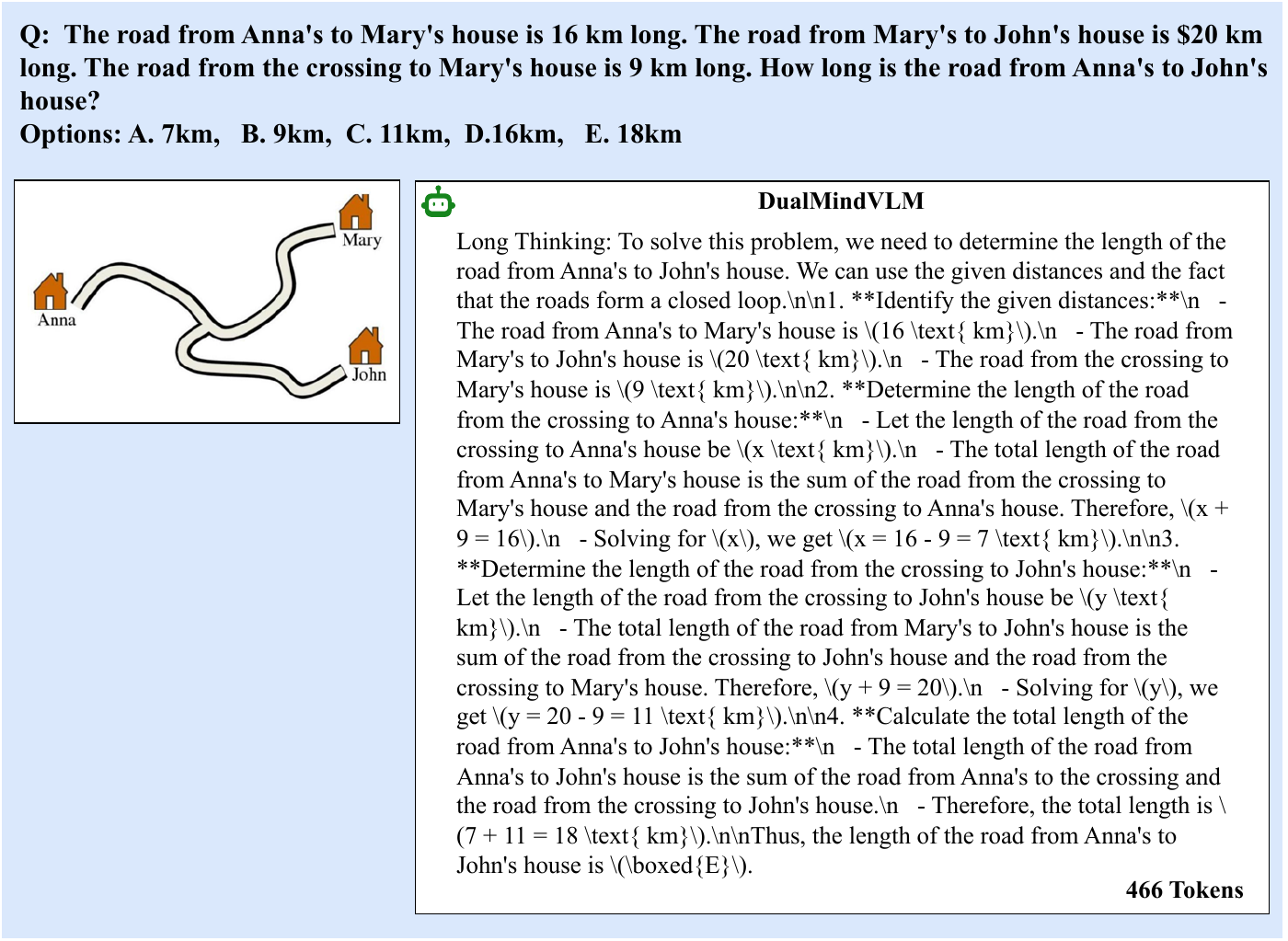}
    \caption{An example response of DualMindVLM to a distance reasoning VQA question.}
    \label{fig:sup_case7}
\end{figure*}

We present case studies illustrating how DualMindVLM adapts to different question types. For relatively simple perception-centric queries (Figures~\ref{fig:sup_case1}–\ref{fig:sup_case4}), the model adopts the fast-thinking mode, reducing token usage while maintaining accuracy compared with GRPO. For more challenging reasoning-oriented queries (Figures~\ref{fig:sup_case5}–\ref{fig:sup_case7}), it switches to the slow-thinking mode, allocating more tokens for detailed step-by-step reasoning.

\clearpage

\clearpage

\clearpage

\clearpage

\clearpage

\bibliography{main}
\end{document}